\documentclass[lettersize,journal]{IEEEtran}
\usepackage{cite}
\usepackage{amsmath,amssymb,amsfonts}
\usepackage{graphicx}
\usepackage{textcomp}
\usepackage{xcolor}
\usepackage[ruled,vlined]{algorithm2e}

\usepackage{algpseudocode}
\usepackage{array}
\usepackage{stfloats}
\usepackage{url}
\usepackage{verbatim}
\usepackage{graphicx}
\usepackage{cite}
\usepackage{mathrsfs}
\usepackage{amsthm}
\usepackage{booktabs}
\usepackage{verbatim}
\usepackage{multirow}
\usepackage{longtable}
\usepackage{booktabs}
\usepackage{makecell}
\usepackage{pifont}
\usepackage{hyperref}
\usepackage[caption=false,font=normalsize,labelfont=sf,textfont=sf]{subfig}
\usepackage{cuted}\stripsep -3pt plus 3pt minus 2pt
\theoremstyle{plain}%设置definition格式
\newtheorem{myDef1}{Definition}
\newtheorem{myDef3}[myDef1]{Definition}%与myDef1同组编号
\newtheorem{myDef4}[myDef1]{Definition}%与myDef1同组编号
\begin{document}

\title{REFOL: Resource-Efficient Federated Online Learning for Traffic Flow Forecasting}

\author{
        Qingxiang Liu,
        Sheng Sun,
        Yuxuan Liang,~\IEEEmembership{Member,~IEEE,}
        Xiaolong Xu,~\IEEEmembership{Senior Member,~IEEE,}
        Min Liu,~\IEEEmembership{Senior Member,~IEEE,}
        Muhammad Bilal,~\IEEEmembership{Senior Member,~IEEE,}
        Yuwei Wang,~\IEEEmembership{Member,~IEEE,}
        Xujing Li,
        and Yu Zheng,~\IEEEmembership{Fellow,~IEEE}
        
        % <-this % stops a space
% \thanks{This work was supported by the National Key Research and Development Program of China (2021YFB2900102) and the National Natural Science Foundation of China (No. 62072436 and No. 62202449).}% <-this % stops a space
\thanks{Qingxiang Liu and Xujing Li are with Institute of Computing Technology, Chinese Academy of Sciences and also with University of Chinese Academy of Sciences, Beijing, China (e-mail: qingxiangliu737@gmail.com and lixujing19b@ict.ac.cn).}
\thanks{Sheng Sun and Yuwei Wang are with Institute of Computing Technology, Chinese Academy of Sciences, Beijing, China (email: sunsheng@ict.ac.cn and ywwang@ict.ac.cn).}
\thanks{Yuxuan Liang is with Intelligent Transportation Thrust, Hong Kong University of Science and Technology (Guangzhou), Guangzhou, China (email: yuxliang@outlook.com).}
\thanks{Xiaolong Xu is with the School of Software, Nanjing University of Information Science and Technology, Nanjing 210044, China, also with the Jiangsu Collaborative Innovation Center of Atmospheric Environment and Equipment Technology (CICAEET), Nanjing University of Information Science and Technology, Nanjing 210044, China (e-mail: njuxlxu@gmail.com).}
\thanks{Min Liu is with Institute of Computing Technology, Chinese Academy of Sciences and also with Zhongguancun Lab, Beijing, China (email: liumin@ict.ac.cn).}
\thanks{Muhammad Bilal is with the School of Computing and Communications, Lancaster University, Bailrigg, Lancaster LA1 4WA, United Kingdom (e-mail: m.bilal@ieee.org).}
\thanks{Yu Zheng is with JD Intelligent Cities Research and JD iCity, JD Technology, Beijing, China (email: msyuzheng@outlook.com).}
\thanks{Min Liu is the corresponding author.}
}

\markboth{REFOL: Resource-Efficient Federated Online Learning for Traffic Flow Forecasting}%
{Shell \MakeLowercase{\textit{et al.}}: A Sample Article Using IEEEtran.cls for IEEE Journals}


\maketitle

\begin{abstract}

Multiple federated learning (FL) methods are proposed for traffic flow forecasting (TFF) to avoid heavy-transmission and privacy-leaking concerns resulting from the disclosure of raw data in centralized methods. However, these FL methods adopt offline learning which may yield subpar performance, when concept drift occurs, i.e., distributions of historical and future data vary. Online learning can detect concept drift during model training, thus more applicable to TFF. Nevertheless, the existing federated online learning method for TFF fails to efficiently solve the concept drift problem and causes tremendous computing and communication overhead. Therefore, we propose a novel method named Resource-Efficient Federated Online Learning (REFOL) for TFF, which guarantees prediction performance in a communication-lightweight and computation-efficient way. Specifically, we design a data-driven client participation mechanism to detect the occurrence of concept drift and determine clients' participation necessity. Subsequently, we propose an adaptive online optimization strategy, which guarantees prediction performance and meanwhile avoids meaningless model updates. Then, a graph convolution-based model aggregation mechanism is designed, aiming to assess participants’ contribution based on spatial correlation without importing extra communication and computing consumption on clients. Finally, we conduct extensive experiments on real-world datasets to demonstrate the superiority of REFOL in terms of prediction improvement and resource economization.

\end{abstract}


\section{Introduction}
\IEEEPARstart{W}{ith} the unprecedented growth of vehicles and unsatisfactory road planning, traffic congestion gets steadily worse and tremendously impacts the macro economy. It is forecast that drivers will expend \$480 \textcolor{black}{billion} for time delay, fuel waste, and carbon emission caused by traffic congestion in the top 25 crowded cities of U.S. by 2027 \cite{afrin2020survey}. 
For alleviating traffic jams, traffic flow forecasting (TFF) is regarded as the foremost component, based on which the administration is capable of foreseeing the burst flow and then taking measures accordingly\cite{guo2019attention}. 
Therefore, TFF has brought about extensive interests both in academia and industry, with great quantity of traffic nodes (i.e., sensors, loop detectors and radar-video integrated machines) deployed along roads generating enormous real-time traffic data \cite{miglani2019deep} and large variety of innovative methods proposed to enhance performance gains in TFF.

Nowadays, most TFF methods rely on powerful deep neural networks to improve prediction performance, with Recurrent Neural Networks (RNN)\cite{8894895} adopted to capture temporal patterns inside traffic flows and Convolutional Neural Networks (CNN) \cite{liang2019urbanfm} as well as Graph Neural Networks (GNN) \cite{zheng2020gman} used for evaluating spatial correlation among traffic nodes to boost performance. 
However, in these \textit{centralized} methods, large volume of historical traffic data generated by traffic nodes should be transmitted to a central server for training prediction models, resulting in considerable transmission overhead.
% What's more, the raw data containing private information of drivers (e.g., location and license plate number) is at high risk of being disclosed during data transmission process.

Federated learning (FL) emerges as a potent solution to tackle these problems above \cite{yang2019federated}, where traffic nodes (termed \textit{clients}) are orchestrated to collaboratively train a global prediction model without disclosure of raw data.
Some recent researches have focused on forecasting traffic flow in FL architecture \cite{liu2020privacy, zhang2021fastgnn, meng2021cross}. 
In \cite{meng2021cross}, clients adopt encoder-decoder architecture of Gated Recurrent Units network (GRU) to model temporal patterns, and the central server uses GNN for updating encoders' hidden states to evaluate spatial correlation among traffic nodes.
However, all of these proposed methods adopt the \textit{offline} learning manner where prediction models are firstly trained based on historical traffic data and then deployed at traffic nodes for future traffic forecasting.
When {concept drift} occurs, i.e., the distribution of historical traffic data and the newly-detected traffic data changes \cite{lu2018learning}, directly applying these pre-trained models to the new traffic data may easily yield unsatisfactory prediction performance.

% Afterwards, these pre-trained models are directly applied to the newly-detected traffic data and easily yield unsatisfactory prediction performance, due to distribution discrepancy between the historical and newly-detected traffic data.
% Therefore, it is bound to re-train and re-deploy the prediction model for performance improvement, unfortunately causing extra resource consumption and evident model update hysteresis.

In contrast, \textit{online} learning (OL) \cite{hoi2021online} makes it possible to detect concept drift upon newly-observed data arriving, and then optimize prediction models accordingly.
With the merits of capturing fluctuating patterns and timely model updates, OL is more applicable for TFF to achieve much more attractive prediction performance.
In \cite{10618965}, Liu \emph{et al.} first propose a Federated Online Learning (FOL) method for traffic flow prediction by integrating OL into FL paradigm.
% However, this FOL method alleviates the problem of performance degradation resulting from concept drift at the sacrifice of enormous computational and communication resources.
Specifically, all clients need to update prediction models once they detect new traffic data, regardless of whether concept drift occurs, which results in large amount of computational cost for local optimization and communication cost for exchanging model parameters.
% However, this FOL method inevitably generates enormous computational and communication overheads due to the need of clients-side online model update and spatial-temporal correlation evaluation at each training round.
% Specifically, temporal patterns frequently change with traffic flow fluctuation, thus making the previous prediction model inapplicable to the newly-arriving traffic data, which is named as concept drift phenomenon \cite{lu2018learning}.
Therefore, it is necessary to reasonably detect concept drift for avoiding resource waste derived from client-side ineffective model updates with little contribution to performance promotion.
% are unbeneficial to performance promotion and simultaneously lead to resource waste.
Besides, it is challenging to design a resource-efficient mechanism for evaluating the time-variant spatial correlation resulting from data fluctuation without frequent client-server parameter transmission and extra computational operations on clients, aiming to decrease resource consumption while guaranteeing prediction performance.

To this end, we propose a novel TFF method in FOL paradigm named Resource-Efficient Federated Online Learning (REFOL), aiming to reduce computational and communication cost at clients while guaranteeing prediction performance. 
% insignificant performance degradation.
% remarkably promote prediction performance with insignificant computing and communication overheads.
% 实现concept drift与开销之间的tradeoff--设计有效的概念漂移检测机制
%高效的空间相关性评估方式
Specifically, we first design a data-driven client participation mechanism, which enables each client to autonomously detect whether concept drift occurs by calculating data distribution divergence and further determine the necessity of local model updates.
Furthermore, clients conduct adaptive online optimization locally to eliminate the influence of concept drift and guarantee performance gains.
Finally, instead of simple averaging mechanism, we adopt a novel graph convolution-based model aggregation mechanism to efficiently evaluate the importance of local updated models depending on time-varying spatial correlation among traffic nodes and further yield the fresh global model with superior generalization ability.
% \textcolor{red}{Subsequently, an adaptive online optimization strategy is proposed to guarantee performance boost and meanwhile reduce communication overheads for exchanging model parameters as well as computing overheads for local optimization.
% Finally, we design a novel graph convolution-based model aggregation strategy, which builds the evaluation of time-varying spatial correlation into model aggregation process with an aim to generate the fresh global model of great generalization and avoid client-side extra resource consumption for evaluating spatial dependence.}
% By introducing this aggregation strategy, the generated fresh global models are of great generalization ability.
The main contributions are summarized as follows:
\begin{itemize}
	\item We propose a novel method called REFOL to precisely forecast traffic flow with spatio-temporal variation in a computation-efficient and communication-lightweight way in federated online learning (FOL) paradigm.
	
	\item We design a data-driven client participation mechanism and an adaptive online optimization strategy, which makes clients detect concept drift and further determine requisite model updates at the least sacrifice of computing and communication resources.
 % We design a data-driven client participation mechanism, which makes clients detect concept drift and determine requisite model update. 
	% Then, we propose an adaptive online optimization strategy to realize performance gains at the least sacrifice of computing and communication overheads.
	
	\item We design a novel graph convolution-based aggregation mechanism to ensure the generalization ability of global models, which can reasonably quantify participants' importance based on spatial correlation evaluation without importing extra resource consumption on traffic nodes. 
	
	\item We validate the efficiency and effectiveness of REFOL by conducting comparison experiments with other FL and FOL prediction methods on two real-world datasets.
 
 % , with performance gains increased by up to 52\% compared with state-of-the-art FL methods.
	
\end{itemize}

The remainder of this paper is organized as below. Section II presents the literature review on \textcolor{black}{TFF}, \textcolor{black}{FL}, and concept drift detection. In Section III, we formulate the TFF problem in FOL paradigm. Section IV elaborates the technical details of the proposed REFOL. 
Then, we analyze the performance of REFOL from experimental results in Section V. Finally, we conclude the paper in Section VI.

\section{Related Work}
%In this section, we introduce the related literature of traffic flow forecasting, federated learning and online learning.

\subsection{\textcolor{black}{TFF}}
Existing TFF methods can be divided into two categories, i.e, parametric and non-parametric methods. The most classical parametric methods are AutoRegressive Integrated Moving Average (ARIMA) \cite{wei2006time} and its variants, e.g., Kohonen-ARIMA (KARIMA) \cite{van1996combining} and subset ARIMA \cite{lee1999application}, 
% and seasonal ARIMA \cite{williams2003modeling}, 
where the parameters are interpretable. 
However, these methods suffer from subpar prediction performance when confronted with complicated traffic flow, since they are based on the assumption that the input sequences are stationary.

\textcolor{black}{With the increasing number of traffic nodes and explosion of traffic data, these parametric methods get less effective and the non-parametric methods based on deep learning models have emerged.}
%Methods of this category based on deep learning have achieved state-of-the-art performance. 
Due to the ability of capturing temporal patterns inside traffic sequences, RNN and its variants, e.g., Long-Short Term Memory network (LSTM) \cite{yang2019traffic}
% \cite{tian2018lstm} 
and GRU \cite{dai2019short}
% \cite{zhang2018combining}
have been adopted in TFF. In \cite{sun2020ssgru}, a novel model called Selected Stacked Gated Recurrent Units (SSGRU) is proposed to make traffic prediction in a road network. In \cite{hsueh2021short}, LSTM is used for short-term TFF, and hidden patterns contained in traffic data are analyzed to increase prediction accuracy.
Furthermore, graph approaches such as GNN, Graph Convolution Network (GCN) and Graph Attention Network (GAT) are integrated to evaluate spatial correlations among traffic nodes.
% \cite{jiang2021graph}. 
\cite{li2018diffusion} proposed Diffusion Convolutional Recurrent Neural Network (DCRNN) and treated traffic flow as a diffusion process. \cite{guo2019attention} proposed the Attention based Spatial-Temporal Graph Convolutional Network (ASTGCN) considering dynamic changes of spatial-temporal correlation. 
Despite performance improvement, these methods adopt centralized training mode, which causes tremendous communication burden and poses a threat on privacy preserving.

\subsection{\textcolor{black}{FL} for \textcolor{black}{TFF}}
Due to the advantages of privacy protection and data localization, some researches focus on forecasting traffic flow in FL paradigm. 
\textcolor{black}{
Depending on what acts as the ``client" in FL, these methods can be divided into three categories.
In the first category, the cities work as clients which can train spatio-temporal prediction models with local datasets. Methods in this category aims to construct the prediction model upon cross-city datasets, thus promising effective prediction model for data-sparse cities by knowledge transfer \cite{zhang2024personalized}.
Methods in the second category try to divide all traffic nodes into multiple organizations \cite{liu2020privacy,xia2022short,zhang2021fastgnn,liu2023multilevel,lou2022stfl,yuan2022fedstn,yang2024fedgtp}. 
In \cite{liu2020privacy,xia2022short,zhang2021fastgnn,liu2023multilevel,lou2022stfl,yuan2022fedstn}, the inter-client dependencies are not considered, hardly guaranteeing the effective evaluation of the spatial correlation.
While in \cite{yang2024fedgtp}, an extra intra-client spatial aggregation module is designed at clients to tackle such challenge.
In the third category, the traffic nodes work as FL clients. 
In \cite{meng2021cross}, traffic nodes are treated as clients and collaboratively update prediction model using local data. 
However, this method requires to transmit hidden states of encoder-GRU from clients to the central server for evaluating spatial correlation, inevitably increasing communication burden and prolonging the occupied time of model updates. 
Furthermore, all of them adopt offline learning manner and cannot meet the need of model re-training driven by data distribution fluctuation.}

Liu \emph{et al.} proposed an Online Spatio-Temporal Correlation-based \textcolor{black}{FL} (FedOSTC) method for traffic flow prediction in \cite{10618965}.
This is the first work on forecasting traffic flow in FOL paradigm.
In FedOSTC, all clients are selected to perform local optimization once they receive new traffic data, in order to mitigate performance decreasing caused by concept drift.
However, this method fails to consider practical model updating requirements on clients and results in unnecessarily tremendous communication and computational overhead.

% \subsection{Online Learning}

% OL mode has the benefits of timely optimizing prediction models with sequentially-coming data to better adapt to traffic flow fluctuation. Thanks to the merits, there have been some works forecasting traffic flow in OL mode. 
% In \cite{6553284}, the authors propose a novel model named Online Learning Weighted Support-Vector Regression (OLWSVR) to forecast short-term traffic flow and show the superiority of the proposed model.
% In \cite{9207661}, OL methods are used for forecasting traffic congestion level. Furthermore, the authors assess dynamic changes of class distribution over time.
% In \cite{9072348}, the authors design an online data-driven model to perform short-term traffic prediction by improving cooperation among roads and vehicles. 
% Although satisfactory performance can be achieved for TFF, all of these methods adopt centralized training mode. 

\subsection{Concept Drift Detection}
%Concept drift means the data distribution changes over time, which makes the existing prediction model unable fit for the upcoming data \cite{lu2016concept,lu2018learning}.
Concept drift reflects the unforeseeable changes in underlying data distribution over time, and consequentially makes the existing prediction model unable to fit the upcoming data, resulting in poor prediction performance \cite{lu2016concept,lu2018learning}. 
It is of great importance to accurately detect concept drift, benefiting for timely model update and efficiently preventing performance degradation.
%Therefore, it is critical to design concept drift detection mechanisms.
Concept drift detection methods can be classified into two categories, i.e, pre-prediction methods and after-prediction methods.
The former devotes to quantifying the distribution divergence between the historical data and new data by certain divergence metrics before conducting prediction \cite{kifer2004detecting,gu2016concept,bu2017incremental}.
The latter first conducts prediction, and then evaluates the prediction errors to judge whether concept drift occurs.
%is mostly based on data distribution. 
%These methods quantify the distribution divergence between the historical data and new data by certain divergence metrics \cite{kifer2004detecting,gu2016concept,bu2017incremental}.
%In after-prediction detection methods, we need to conduct prediction first and then evaluate the evaluate the prediction error rate. 
%If the online error rate reaches a preset threshold, the  prediction model will be replaced. 
The representative methods are Drift Detection Drift (DDM) \cite{gama2004learning}, Early DDM (EDDM) \cite{baena2006early}, Statistical Test of Equal Proportions Detection (STEPD) \cite{nishida2007detecting}, \emph{etc.}

\section{Preliminaries and Problem Formulation}
In this section, we present the introduction of \textcolor{black}{TFF} and concept drift detection. Then we formulate the problem of \textcolor{black}{TFF} in FOL paradigm. Finally, we present a vanilla FOL method for predicting traffic flow.
For better understanding, we list primary symbols and their descriptions in Table \ref{symbol}.

\begin{table}
    \caption{Primary Symbols and Descriptions in Section III}
  \small
\begin{tabular}{ll}
    
    \toprule
    \textbf{Symbol} & \textbf{Description} \\
    \midrule
    $r_n$ & The $n$-th traffic node.\\
    $N$ & Number of traffic nodes.\\
    $e_{m,n}$ & Adjacent edge between $r_n$ and $r_m$.\\
    $\mathcal{N}$ & Traffic node set.\\
    $\mathcal{E}$ & Edge set among traffic nodes.\\
    $\mathcal{G}$ & Transportation network.\\
    $s_{t,n}$ & Traffic data detected by $r_n$ at the $t$-th round.\\
    $\mathcal{T}$ & Number of rounds.\\
    $H$ & Historical horizon.\\
    $F$ & Forecasting horizon.\\
    $hw_n$ & Historical time window of $r_n$.\\
    $fw_n$ & Future time window of $r_n$.\\
    $f(\cdot)$ & Prediction model.\\
    $S_{t,n}^H$ & Inputting data of $r_n$ at the $t$-th round.\\
    $\hat{s}_{t,n}$ & Predicted value of $s_{t,n}$.\\
    $\hat{S}_{t,n}^F$ & Predicted values of $r_n$ at the $t$-th round.\\
    $S_{t,n}^F$ & True values of $r_n$ at the $t$-th round.\\
    $w_{t,n}$ & Model parameters of $r_n$ at the $t$-th round.\\
    $\mathcal{L}(\cdot)$ & Loss function.\\
    $w_{n}^{*}$ & Local optimal model of $r_n$.\\
    $rgt_n$ & Regret of $r_n$.\\
    $w^*$ & Global optimal model.\\
    \bottomrule
\end{tabular}
\label{symbol}
\end{table}

% In this section, we firstly formulate the traffic flow forecasting problem in federated online learning paradigm, and then elaborate the implementation of FL with OL mode for the formulated problem. 
\subsection{\textcolor{black}{TFF}}
A traffic node is deployed at a certain road segment and responsible for monitoring the traffic speed of this segment. 
These traffic nodes compose a \textit{transportation network}, which can be represented as a directed graph $\mathcal{G}=\left(\mathcal{N},\mathcal{E}\right)$. $\mathcal{N}=\left\{ {{r_n}\left| {1 \le n \le N} \right.} \right\}$ denotes the traffic node set, and $r_n$ represents the $n$-th traffic node. $\mathcal{E}=\left\{ {{e_{m,n}}\left| {1 \le m,n \le N} \right.} \right\}$ denotes the edge set. If $r_n$ is adjacent to $r_m$, $e_{m,n} \in \mathcal{E}$. Otherwise, $e_{m,n} \notin\mathcal{E} $. 
Let $\mathcal{S} = \left\{\mathcal{S}_1, \cdots, \mathcal{S}_t, \cdots, \mathcal{S}_{\mathcal{T}}\right\}$ denote the traffic data set, where $\mathcal{T}$ denotes the total number of time stamps and $\mathcal{S}_t = \left[ s_{t,n}\right]_{1\le n\le N}$. $s_{t,n} \in \mathbb{R}^1$ represents the traffic speed detected by $r_n$ at the $t$-th time stamp.
The \textcolor{black}{TFF} problem can be viewed as a process of predicting future traffic speed based on historical and current traffic speed. 
Without loss of generality, we consider to perform traffic prediction of $F$ forecasting steps at the $t$-th time stamp based on the latest $H$ historical steps, which can be formulated as $\left\{ \mathcal{S}_{t-H+1},  \cdots , \mathcal{S}_{t};\mathcal{G} \right\} \stackrel{\mathcal{F} }{\longrightarrow}
    \left\{ \hat{\mathcal{S}}_{t+1},  \cdots, \hat{\mathcal{S}}_{t+F} \right\}$.
$\hat{\mathcal{S}}_{t+\delta} (1\le \delta\le F)$ denotes the predicted value of $\mathcal{S}_{t+\delta}$ and $\mathcal{F}$ represents the prediction algorithm.
The above process of forecasting traffic flow actually tackles the problem of modeling the spatio-temporal correlation, i.e., how to capture temporal dependence inside each traffic flow and evaluate spatial correlation among different traffic flows.

% \subsection{Concept Drift Detection}
% The traffic speed is bound to be affected by many factors such as weather condition and emergency. 
% Therefore, traffic flows to be forecasted can change over time in an unforeseen way and make the pre-trained prediction model inefficacious, which is named as \textit{concept drift} \cite{lu2018learning}.
% The aim of concept drift detection is to find whether and when concept drift occurs. Given the historical time window $hw_n$ and future time window $fw_n$ for $r_n$. The distribution difference between traffic data in $hw_n$ and $fw_n$ is evaluated. 
% If the distribution changes, concept drift occurs at this time stamp. $r_n$ sets $hw_n = fw_n$, and slides $fw_n$ backwards by one step. Otherwise, $r_n$ keeps $hw_n$ fixed and $fw_n$ slides one step backwards.

\subsection{FOL}
%The centralized prediction methods for solving the problem in (1) require traffic nodes uploading their detected traffic data to the central server for training the prediction models.
In FOL paradigm, each traffic node is regarded as a \textit{client}. 
Once detecting new traffic data, the clients adopt the locally-saved prediction models or request the fresh global model for prediction, and perform local update if necessary.
The central server aggregates updated local models.
At the $t$-th time stamp (i.e., $t$-th round in FOL paradigm), the prediction problem of $r_n$ can be formulated as
\begin{equation}
    \hat S_{t,n}^F = f\left(S_{t,n}^H;{w_{t,n}} \right),
\end{equation}
where $f(\cdot)$ denotes the adopted prediction model and $w_{t, n}$ denotes the model parameters of $r_n$ at the $t$-th round. 
$S_{t,n}^H=\left({s_{t - H + 1,n}},{s_{t - H + 2,n}},...,{s_{t,n}}\right)$ stands for the input of the prediction model, and $\hat S_{t,n}^F = \left({\hat s_{t + 1,n}},{\hat s_{t + 2,n}},...,{\hat s_{t + F,n}}\right)$ stands for the predicted value of $S_{t,n}^F = \left( {{s_{t + 1,n}},{s_{t + 2,n}}, \cdots ,{s_{t + F ,n}}} \right)$, where ${\hat s_{t + 1,n}}$ denotes the predicted value of $s_{t+1,n}$. 

\begin{myDef3}[\textbf{Local Optimal Model}]
    \textcolor{black}{We assume that $r_n$ performs prediction distributedly.}
    After $\mathcal{T}$ rounds, all traffic data $\left\{ s_{1,n}, s_{2,n}, \cdots, s_{\mathcal{T},n}  \right\}$ are available to $r_n$. Therefore, the integrated training dataset $ \mathcal{TD}_{n} =  \left\{ {\left( {S_{t,n}^H,S_{t,n}^F} \right)\left| {1 \le t \le \mathcal{T}} \right.} \right\}$ could be obtained. 
    $r_n$ could optimize the prediction model $f(\cdot)$ based on $\mathcal{TD}_n$ and would obtain the local optimal model, which is denoted as $w_n^{\ast}$. Formally, $w_n^{\ast}$ is formulated as
    \begin{equation}
        w_n^ *  = \mathop {\arg \min }\limits_w \left\{ {\sum\nolimits_{t = 1}^\mathcal{T} {\mathcal{L}\left(f\left(S_{t,n}^H;w\right),S_{t,n}^F\right)} } \right\},
    \end{equation}
    where $\mathcal{L}(\cdot)$ stands for the loss function to evaluate the discrepancy between the predicted and true values. 
    It is intuitive that for the given dataset $\mathcal{TD}_n$ and the prediction model $f(\cdot)$, the local optimal model is fixed.
\end{myDef3}

We define $rgt_n$ as the prediction regret of $r_n$, which is obtained via evaluating the prediction loss difference between the actual prediction model $w_{t,n} (1\le t\le \mathcal{T})$ and the local optimal model $w_n^{\ast}$ over all $\mathcal{T}$ rounds. 
The calculation of $rgt_n$ with respect to $w_n^*$ can be expressed as
\begin{align}
rgt_n(w^*_n) & = \sum\nolimits_{t = 1}^{\mathcal{T}} {\mathcal{L}(f(S_{t,n}^H;w_{t,n}),S_{t,n}^F)}
\notag
\\
&-\sum\nolimits_{t = 1}^{\mathcal{T}} {\mathcal{L}(f(S_{t,n}^H;{w_n^*}),S_{t,n}^F)}.
\end{align}

\begin{myDef4}[\textbf{Global Optimal Model}]
    We denote $\mathcal{TD} = \left\{ \mathcal{TD}_n | {1\le n\le N } \right\}$ as the integrated training datasets of all clients. If we input $\mathcal{TD}$ into $f(\cdot)$ and perform optimization, the global optimal model $w^*$ can be obtained, which is formulated formally as
    \begin{equation}
        {w^ * } = \mathop {\arg \min }\limits_w \left\{ {\sum\nolimits_{n = 1}^N {\sum\nolimits_{t = 1}^\mathcal{T} {\mathcal{L}(f(S_{t,n}^H;w),S_{t,n}^F)} } } \right\}.
    \end{equation}
    For the given $\mathcal{TD}$ and $f(\cdot)$, the global optimal model is also fixed.
\end{myDef4}

The objective of \textcolor{black}{TFF} in FOL paradigm is to minimize the prediction regret with respective to the global optimal model over all $N$ traffic nodes, which is formulated as
\begin{equation}
{ \min } \left\{ {\sum\nolimits_{n = 1}^N {rg{t_n}(w^{*})} } \right\}.
\end{equation}
\begin{align}
{\rm where}\ \  rgt_n(w^*) &= \sum\nolimits_{t = 1}^{\mathcal{T}} {\mathcal{L}(f(S_{t,n}^H;w_{t,n}),S_{t,n}^F)}
\notag
\\
&-\sum\nolimits_{t = 1}^{\mathcal{T}} {\mathcal{L}(f(S_{t,n}^H;{w^*}),S_{t,n}^F)}.\notag
\end{align}

% \subsection{Preliminary Knowledge of Federated Learning with Online Learning Manner}
\subsection{Vanilla FOL Method}
Federated Averaging (FedAvg) is a vanilla \textcolor{black}{FL} method which adopts offline learning \cite{mcmahan2017communication}. 
We integrate OL into FedAvg for forecasting traffic flow, and the execution process is elaborated in Algorithm \ref{ag1}. 
Specifically, the central server first randomly selects a subset of available clients denoted as $\mathcal{N}_t$ to participate in this round.
Each selected client concurrently makes prediction (Line 9) and updates model parameters via Online Gradient Descent (OGD) \cite{hoi2021online} for $E$ epochs (Line 10-11).
After finishing online optimization locally, these selected clients transmit updated model parameters to the server, which conducts averaging aggregation to generate the fresh global model (Line 6).
The process continues until no new traffic data arrive.

\begin{algorithm}
	\caption{FOL-vanilla}
	\label{ag1}
	\LinesNumbered
	\KwIn{Initialized model $w_1$, learning rate $\eta$.}
	\KwOut{The global model $w_{\mathcal{T}+1}$.}
	
	\SetKwFunction{Fexecute}{ClientExecute}
	\textsc{\textbf{ServerExecute:}}\\
	\For {$t = 1, 2, \cdots, \mathcal{T}$}{
		$\mathcal{N}_t$ $\gets$ randomly select $N_t$ clients.\\
		\For {$r_n$ in $\mathcal{N}_t$}{
			{$w_{t+1,n} \gets $ \Fexecute($w_t$, $n$)}\\
			
		}
		{${w_{t+1}} \gets \frac{1}{{{N_t}}}\sum\nolimits_{{r_n} \in {\mathcal{N}_t}} {{w_{t+1,n}}} $}\\
	}
	\KwRet $w_{\mathcal{T}+1}$\\
	
	\SetKwProg{Fn}{Function}{:}{}
	\Fn{\Fexecute{$w$, $n$}}{
        Make prediction via Eq. (1).\\
		\For{$e = 1, 2, \cdots, E$}{
			
			$w \gets w - \eta \nabla \mathcal{L}(f(S_{t,n}^H;{w}),S_{t,n}^F)$\\
		}
		\KwRet $w$}
\end{algorithm}

\section{Methodology}
\begin{figure*}[!htbp]
	\centering
	\includegraphics[width=1.0\textwidth]{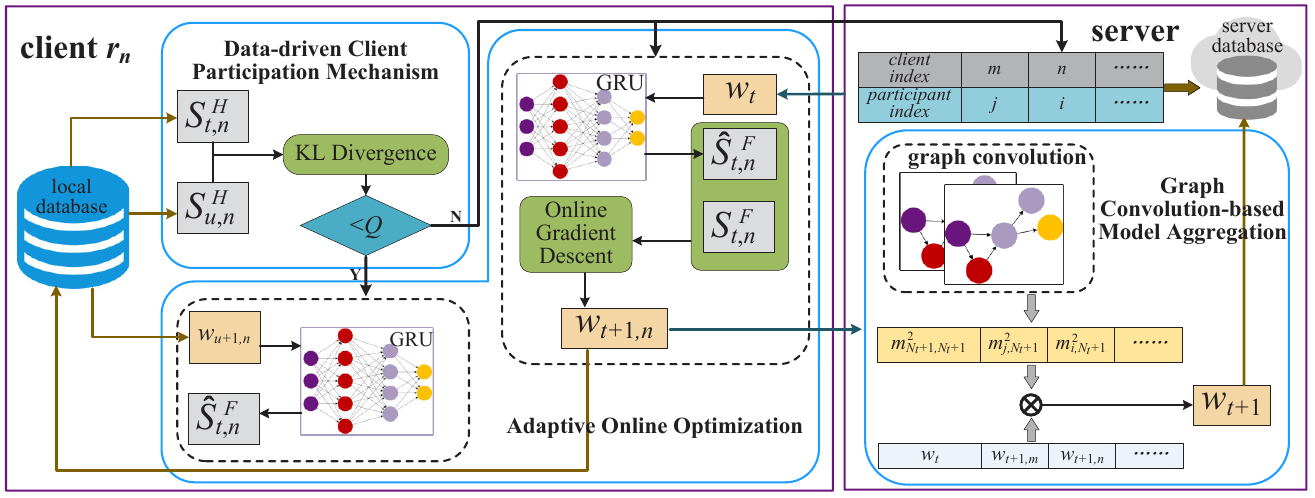}
	\caption{The architecture of REFOL includes three modules, i.e., {data-driven participation mechanism}, {adaptive online optimization}, and {graph convolution-based model aggregation}. Each client determines whether to participate in this round of training based on data-driven client participation mechanism and further performs adaptive online optimization accordingly. The central server collects local optimized model parameters from participants and conducts graph convolution-based model aggregation.}
	\label{method}
\end{figure*}

In this section, we elaborate the proposed \textcolor{black}{TFF} method, i.e, resource-efficient federated online learning (REFOL).
The architecture of REFOL is shown in Fig. \ref{method}. Specifically, REFOL is composed of \emph{data-driven client participation mechanism}, \emph{adaptive online optimization}, and \emph{graph convolution-based model aggregation}.
% We give the detailed introduction of these three components and finally present the algorithm summary of REFOL.

\subsection{Data-driven Client Participation Mechanism}
% Similar to traditional FL paradigm, in FOL paradigm, the server also needs to select a subset of clients to participate in each round of training. 
% In addition, raw traffic data are kept at clients, which are not accessible by the aggregator. 
% Therefore, clients themselves, rather than the aggregator, should detect the occurrence of concept drift for the sake of online prediction performance improvement.

Intuitively, if the traffic flow distribution changes (i.e., \textit{concept drift} occurs), the model parameters need to be updated to guarantee the prediction performance.
It is a straightforward solution that in each round, the server selects all clients as participants to timely optimize their local prediction models based on input data.
However, some clients do not need to perform optimization, if their historically-saved prediction models still work. 
Hence, this simple approach inevitably generates needless communication overhead for exchanging model parameters and compute cost for local optimization.
To this end, we design a data-driven client participation mechanism which can skillfully detect the occurrence of concept drift based on traffic data distribution, and further offer clients the autonomy to decide whether to participate in local training.

% Suppose at the $(\tau +1)$-th round, each client $r_n$ firstly detects traffic speed $s_{\tau+1,n}$ and generates traffic sequence $S_{\tau+1,n}^H$. 
% %It then executes the following processes.
% We first evaluate whether the locally-saved prediction model $w_{\tau +1,n}$ can still yield comparable prediction result via comparing distribution difference between $S_{\tau ,n}^H$ and $S_{\tau +1,n}^H$.
% If the distribution variation is insignificant, it means concept drift doesn't occur at the $\tau+1$-th round on $r_n$, and $r_n$ will reuse $w_{\tau +1,n}$ for traffic prediction. 
% In this case, $r_n$ doesn't need to download the current global model $w_{\tau+1}$ from the central server and perform local optimization, conductive to save communication overhead for exchanging model parameters and computational overhead for local optimization.
% Otherwise, $r_n$ has to download $w_{\tau+1}$ for subsequent prediction and then update local model with $S_{\tau+1,n}^H$.
% Therefore, it is pivotal to evaluate the applicability of $w_{\tau+1,n}$, which can be transformed as concept drift detection at the $\tau +1$-th round.
% Next, we illustrate the process of concept drift detection.

\begin{figure}[!ht]
	\centering
	\includegraphics[width=0.48\textwidth]{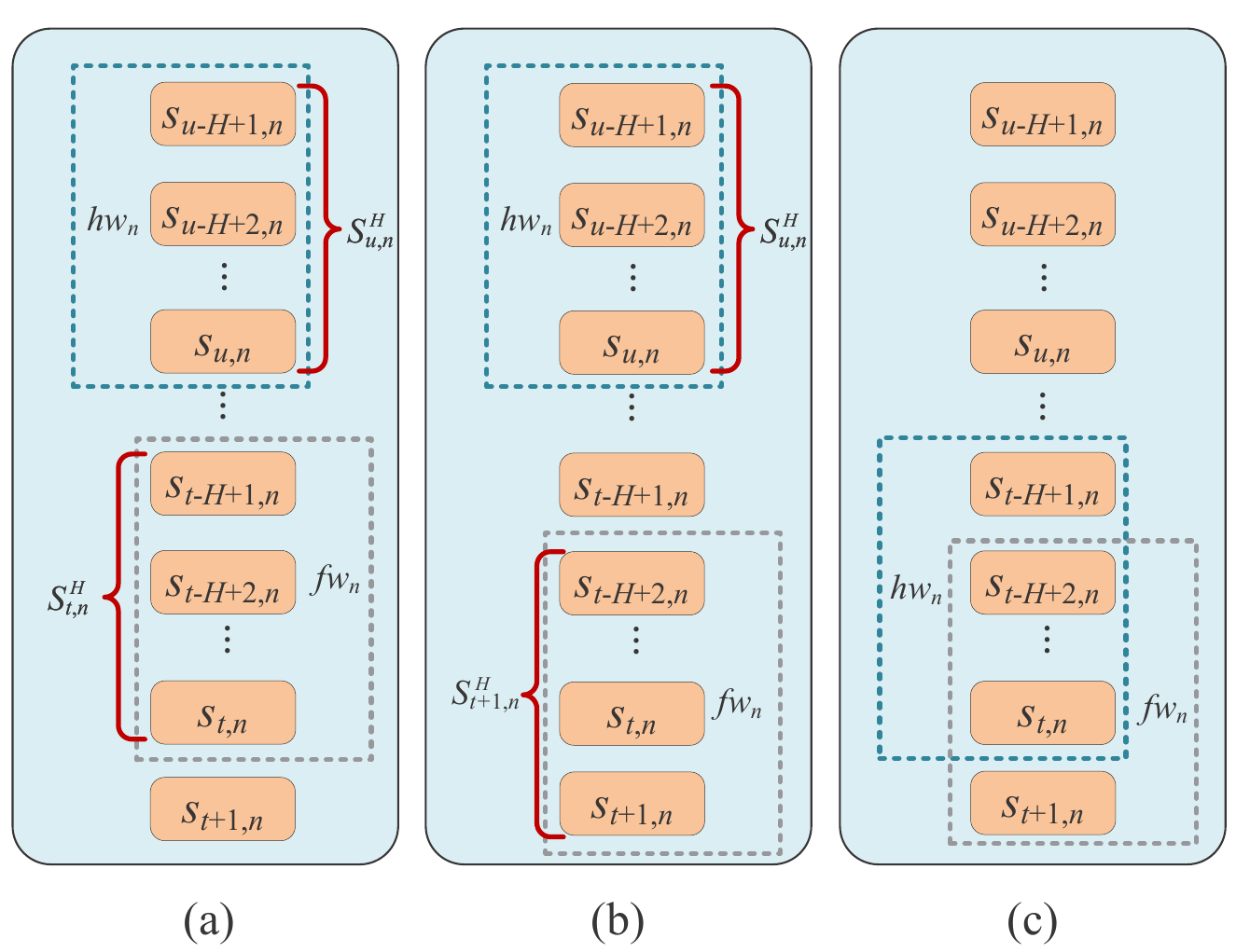}
	\caption{The changing process of $hw_n$ and $fw_n$ in concept drift detection.}
	\label{sliding_window}
\end{figure}

Specifically, assuming the locally-saved model of $r_n$ is generated at the $u$-th round based on traffic sequence $S_{u,n}^H$, denoted as $w_{u+1,n}$.
At the $t$-th round ($u+1\le t\le \mathcal{T}$), we set $hw_n = S_{u,n}^H$ and $fw_n = S_{t,n}^H$, as is shown in Fig. \ref{sliding_window}(a).
Then, we utilize widely-used Kullback-Leibler Divergence (KLD) \cite{joyce2011kullback} to calculate the distribution divergence between $S_{t,n}^H$ and $S_{u,n}^H$, which can be denoted as ${D_{KL}}( S_{t,n}^H|| S_{u,n}^H)$. Higher KLD value indicates the distribution of $S_{t,n}^H$ is less similar to $S_{u,n}^H$, and vice verse.

% Firstly, $S_{t,n}^H$ is normalized to $\bar S_{t,n}^H$ as
% \begin{align}
% \bar S_{t,n}^H &= \left( {\frac{{{s_{t - H + 1,n}}}}{{\sum\nolimits_{i = 0}^{H - 1} {{s_{t - i,n}}} }},\frac{{{s_{t - H + 2,n}}}}{{\sum\nolimits_{i = 0}^{H - 1} {{s_{t - i,n}}} }},...,\frac{{{s_{t,n}}}}{{\sum\nolimits_{i = 0}^{H - 1} {{s_{t - i,n}}} }}} \right)
% \notag\\
% &= \left( {{{\bar s}_{t - H + 1,n}},{{\bar s}_{t - H + 2,n}},...,{{\bar s}_{t,n}}} \right).
% \end{align}

% Similarly, $S_{u,n}^H$ is also normalized to $\bar S_{u,n}^H$.
% Furthermore, we define ${D_{KL}}(\bar S_{t,n}^H||\bar S_{u,n}^H)$ as the divergence between $\bar S_{t,n}^H$ and $\bar S_{u,n}^H$, which is calculated as
% \begin{equation}
% {D_{KL}}(\bar S_{t,n}^H||\bar S_{u,n}^H) = \sum\limits_{i=0}^{H-1} {{{\bar s}_{t-i,n}}} \ln \frac{{{{\bar s}_{t-i,n}}}}{{{{\bar s}_{u-i,n}}}}.
% \end{equation}
% The value of ${D_{KL}}(\bar S_{t,n}^H||\bar S_{u,n}^H)$ can represent the distribution similarity between $S_{t,n}^H$ and $S_{u,n}^H$. 
% Higher KLD value indicates the distribution of $S_{t,n}^H$ is less similar to $S_{u,n}^H$, and vice verse.

We define $Q$ as the threshold of KLD values. 
If ${D_{KL}}$ doesn't reach $Q$, we draw the conclusion that $w_{u+1,n}$ is still competent on the current traffic flow $S_{t,n}^H$. 
Therefore, $r_n$ does not need to download the up-to-date global model from the central server and then perform local model updates.
In this case, $hw_n$ keeps stable, while $fw_n$ moves one step backwards, as is shown in Fig. \ref{sliding_window}(b).
Otherwise, we regard that concept drift occurs at the $t$-th round and $w_{u+1,n}$ is no longer applicable to $S_{t,n}^H$. Therefore, $r_n$ should participate in the $t$-th round federated training (details in the next subsection).
In this case, the historical window is updated as $hw_n = S_{t,n}^F$, as is shown in Fig. \ref{sliding_window}(c).
% and adopts the downloaded global model $w_{t}$ for forecasting in case of prediction performance degradation.
%$r_n$ should request the fresh global model $w_{t-1}$ from the central server. Namely, $r_n$ is selected to participate in the $t$-th round.

%For the central server, upon receiving the request from $r_n$, it is going to distribute $w_{t-1}$ to $r_n$. Otherwise, $r_n$ will have the risk of prediction performance degradation, since $w_{u,n}$ is no longer applicable to $S_{t,n}^H$.

\subsection{Adaptive Online Optimization}
Each client firstly determines whether to participate in this round of training based on the subsection above.
% and then preforms prediction.
To improve prediction performance with least communication and computing overhead, we design an adaptive online optimization strategy, which is described from the following two cases.

\textbf{The first case:} When ${D_{KL}}( S_{t,n}^H|| S_{u,n}^H) < Q$, $r_n$ regards non-occurrence of concept drift and will not participate in the $t$-th round. It will utilize the previous local model to perform forecasting operation (i.e., $w_{t,n}=w_{u+1,n}$), which can avoid unnecessary communication overhead for downloading the up-to-date global model.

We reformulate $S_{t,n}^H$ as $S_{t,n}^H=(s_{a,n}, t-H+1\leq a \leq t)$. 
% After detecting traffic data $s_{t,n}$, 
% $r_n$ utilizes the current prediction model $w_{t,n}$ to online forecast the future traffic speed with inputting traffic sequence $S_{t,n}^H=(s_{a,n}, t-H+1\leq a \leq t)$.
In this paper, we adopt the widely-used GRU model \cite{cho-etal-2014-learning} as the prediction model.
Concretely, the hidden state $h_{n}^a$ can be obtained as follows:
\begin{equation}
{u_n^a} = \sigma \left( {{W_n^{(u)}}{s_{a,n}} + {U_n^{(u)}}{h_n^{a - 1}}} \right),
\end{equation}
\begin{equation}
{r_n^a} = \sigma ({W_n^{(r)}}{s_{a,n}} + {U_n^{(r)}}{h_n^{a - 1}}),
\end{equation}
\begin{equation}
{{h}_n^{a'}} = \tanh (W_n^{(h)}{s_{a,n}} + {r_n^a} \odot U_n^{(h)}{h_n^{a - 1}}),
\end{equation}
\begin{equation}
{h_n^a} = {u_n^a} \odot {h_n^{a - 1}} + (1 - {u_n^a}) \odot {{h}_n^{a'}},
\end{equation}
where $W_n^{(u)} \in \mathbb{R}^{1\times hs}, U_n^{(u)} \in \mathbb{R}^{hs \times hs}, W_n^{(r)}\in \mathbb{R}^{1\times hs}, U_n^{(r)}\in \mathbb{R}^{hs \times hs}, W_n^{(h)}\in \mathbb{R}^{1\times hs}$ and $U_n^{(h)}\in \mathbb{R}^{hs \times hs}$ are the parameters of the GRU network. $hs$ denotes hidden size of the GRU network.
$\sigma (\cdot)$ represents the sigmoid function. 
$u_n^a$ and $r_n^a$ denote the update gate and reset gate respectively. 
The above-mentioned process is iterated for $H$ times and then the hidden states $\{h_n^a | t-H+1\leq a \leq t\}$ are fed into a fully connected layer for yielding the final predicted value $\hat S_{t,n}^F$. 
% It is persuasive that the prediction performance is satisfactory due to the usability of $w_{u+1,n}$. 

Since temporal patterns of local traffic data remain stable, $r_n$ has no effects on changes in spatial correlation and fails to contribute to the fresh global model.
Therefore, $r_n$ will neither locally optimize $w_{t,n}$ nor upload $w_{t,n}$ to the central server for aggregation, thus further saving computing and communication resources.

\textbf{The second case:} When ${D_{KL}}( S_{t,n}^H|| S_{u,n}^H) \ge Q$, $r_n$ regards concept drift occurs and hence the locally-saved model $w_{u+1,n}$ is not applicable to $S_{t,n}^H$. Therefore, $r_n$ will participate in the $t$-th round. We name $r_n$ a \textit{participant}. It first downloads the global model $w_{t}$ as the prediction model, i.e., $w_{t,n}=w_t$, and then generates the predicted value $\hat S_{t,n}^F$ with the same operations as Eq. (6)-(9).
After performing prediction, $r_n$ starts to optimize its local model parameters via OGD \cite{hoi2021online} for $E$ epochs, which can be formulated as
\begin{equation}
w_{t,n}^{(e)} = w_{t,n}^{(e - 1)} - \eta \nabla \mathcal{L}(f(S_{t,n}^H;w_{t,n}^{(e - 1)}),S_{t,n}^F),
\end{equation}
where $w_{t,n}^{(0)}=w_{t,n}$ and $1 \le e \leq E$.  
After local optimization, $r_n$ saves $w_{t+1,n}$ ($w_{t+1,n}=w_{t,n}^{(E)}$) as its local model, and then uploads $w_{t+1,n}$ to the central server for aggregation.

The execution process of data-driven client participation mechanism and adaptive online optimization is summarized in Algorithm \ref{ag2}.
% Firstly, $r_n$ normalizes $S_{t,n}^H$ and $S_{u,n}^H$ respectively and then evaluates distribution divergence (Line 1, 2). 
If the KLD value ${D_{KL}}(\bar S_{t,n}^H||\bar S_{u,n}^H) < Q$, the locally-saved prediction model $w_{u+1,n}$ still works and $r_n$ directly performs prediction (Line 2-4).
Otherwise, $r_n$ considers the occurrence of concept drift at the $t$-th time stamp and downloads the up-to-date global model $w_t$ for prediction and then performs local optimization (Line 5-11).
$r_n$ saves the locally-optimized model $w_{t+1,n}$ as well as traffic sequence $S_{t,n}^H$, and pops out $w_{u+1,n}$ and $S_{u,n}^H$ from local database.
\begin{algorithm}
	\caption{Data-driven Client Participation Mechanism and Adaptive Online Optimization}
	\label{ag2}
	\LinesNumbered
	\KwIn{Locally-saved model $w_{u+1,n}$, $S_{t,n}^H$, $S_{u,n}^H$, global model $w_t$, and learning rate $\eta$.}
	\KwOut{Predicted value $\hat{S}_{t,n}^F$.}
        
        Calculate divergence ${D_{KL}}( S_{t,n}^H|| S_{u,n}^H)$.\\
        \If{${D_{KL}}( S_{t,n}^H|| S_{u,n}^H) < Q$}{
            $w_{t,n} \gets w_{u+1,n}$\\
            \textcolor{black}{$r_n$ performs prediction via Eq. (6)-(9) and generates predicted value $\hat{S}_{t,n}^F$.} \\
            % \KwRet $\hat{S}_{t,n}^F$, $w_{u+1,n}$, and $S_{u,n}^H$.\\
        }
        \Else{
            $r_n$ downloads the global model $w_{t}$.\\
            $w_{t,n} \gets w_t$\\
            \textcolor{black}{$r_n$ performs prediction via Eq. (6)-(9) and generates predicted value $\hat{S}_{t,n}^F$.}\\
            $w_{t,n}^{(0)} \gets w_{t,n}$\\
            \For {$e \le E$}{
			Perform local optimization as Eq. (10).\\
		}
            $w_{t+1,n} \gets w_{t,n}^{(E)}$\\
            Keep $w_{t+1,n}$ locally.\\
            Upload $w_{t+1,n}$ to the aggregator.\\
            % \KwRet $\hat{S}_{t,n}^F$, $w_{t+1,n}$, and $S_{t,n}^H$.
        }

\end{algorithm}

\subsection{Graph Convolution-based Model Aggregation}

\begin{figure*}[!ht]
	\centering
	\includegraphics[width=0.98\textwidth]{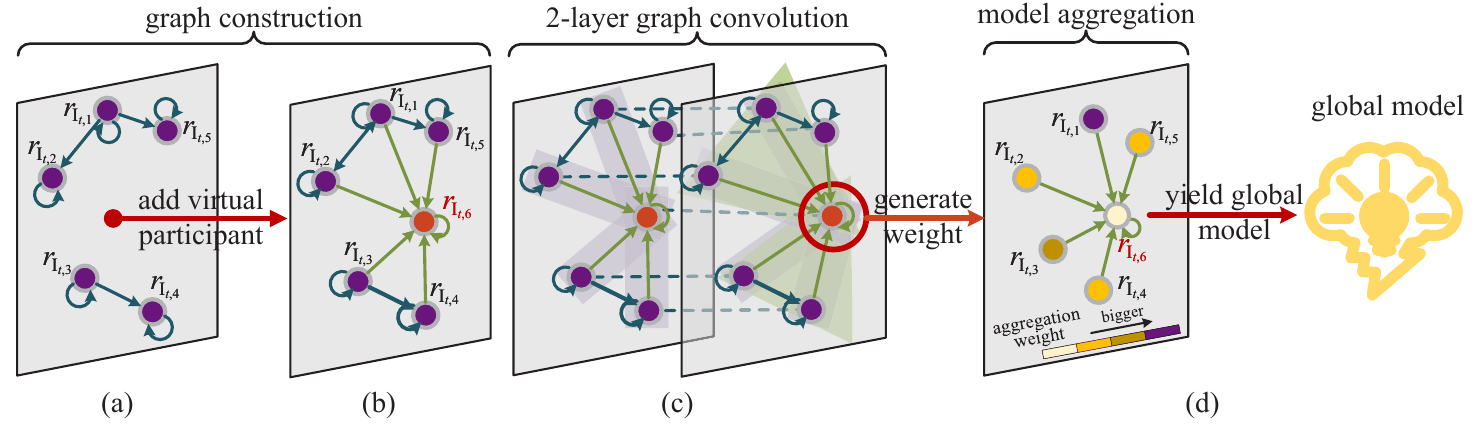}
	\caption{The execution process of graph convolution-based model aggregation contains three parts, i.e., graph construction, 2-layer graph convolution , and model aggregation.}
	\label{aggregation}
\end{figure*}

% Accordingly, if ${D_{KL}}(\bar S_{t,n}^H||\bar S_{u,n}^H) \ge Q$, $r_n$ will participate in the $t$-th round and is regarded as a \textit{participant}.
% Before we elaborate graph convolution-based model aggregation, we first give the definition of \textit{participant}.
% \begin{myDef5}[\textbf{Participant}]
%     At the $t$-th round, according to data-driven client participation mechanism, 
%     $r_n$ has to download the fresh global model and then perform local optimization if ${D_{KL}}(\bar S_{t,n}^H||\bar S_{u,n}^H) \ge Q$. Therefore, $r_n$ becomes a participant at the $t$-th round.
% \end{myDef5}
Generally, different participants exhibit diverse importance in TFF scenarios due to the coupling spatial correlation.
Therefore, it is necessary to govern aggregation weights of participating nodes based on sophisticated spatial correlation for obtaining the updated global model with satisfactory generalization ability. 
However, existing FL methods  evaluate participants' importance by importing an additional assessment procedure of spatial correlation, calling for frequent transmission and thus resulting in tremendous communication overhead.
%and then the central server updates the hidden states by GNN, and returns the refreshed ones to corresponding participants. 
%Furthermore, the backpropagation of GNN needs gradient communication between participants and the central server. 
Thus, we design a novel graph convolution-based model aggregation mechanism, the execution process of which is shown in Fig. \ref{aggregation}.
It leverages graph convolution to online quantify time-variant spatial correlation among participants and correspondingly yields aggregation weights in a communication-lightweight and computation-efficient manner.
The whole procedure is elaborated as follows.

%Compared with existing methods, this mechanism embodies the evaluation of spatial correlation based graph convolution, which is lightweight in terms of computation and avoids transmitting extra parameters from participants.

\subsubsection{\textbf{Graph Construction}}
Suppose $N_t$ clients participate in the $t$-th round according to the designed data-driven client participation mechanism.
These participants execute the process of adaptive online optimization and subsequently upload the updated local models to the central server for generating the fresh global model $w_{t+1}$. 
%We might as well name these clients as ``contributors", since they contribute to the global model of the next round.
They can also compose a directed graph, denoted as $\mathcal{G}_t=(\mathcal{N}_t, \mathcal{E}_t)$. 
$\mathcal{N}_t = \left\{ r_{{\rm I}_{t,i}} | 1\le i\le N_t \right\}$, where ${\rm I}_{t,i}$ represents the client index of the $i$-th participant at the $t$-th round and  $\mathcal{N}_t \subset \mathcal{N}$.
As is shown in Fig. \ref{aggregation}(a), the purple nodes represent the participants.
$\mathcal{E}_t = \left\{ e_{{\rm I}_{t,i}, {\rm I}_{t,j}} | 
e_{{\rm I}_{t,i}, {\rm I}_{t,j}} \in \mathcal{E} \right\}$ represents the edge set of these participants.
It is explicit that $\mathcal{E}_t \subset \mathcal{E}$ and $\mathcal{G}_t$ is a subgraph of $\mathcal{G}$. Let ${A_t}=\left[a_{i,j} \right]_{1\le i,j\le N_t}$ denote the adjacent matrix. If ${e_{{{\rm I}_{t,i}},{{\rm I}_{t,j}}}} \in {\mathcal{E}_t}$, $a_{i,j} =1 $. Otherwise $a_{i,j} =0$.
Let ${D_t}= diag \left( {d_{{\rm I}_{t,1}}}, \cdots, {d_{{\rm I}_{t,i}}}, \cdots, {d_{{\rm I}_{t,N_{t}}}} \right)$ denote the indegree matrix, where ${d_{{\rm I}_{t,i}}}$ represents the indegree of $r_{{\rm I}_{t,i}}$. 

We treat local models as participants' features and perform graph convolution on $\mathcal{G}_t$. 
If $\mathcal{G}_t$ is not a connected graph (as is shown in Fig. 3(a)), no participants will aggregate features (local models) from all the other ones after graph convolution.
%Therefore, each participant will aggregate the local models of its adjacent ones.
In order to obtain the up-to-date global model after graph convolution, we introduce a virtual participant $r_{{\rm I}_{t,N_t +1}}$ (the red node in Fig. \ref{aggregation}(b)), and $w_{t}$ is regarded as its feature. 
Furthermore, we define that all participants are adjacent to $r_{{\rm I}_{t,N_t +1}}$ and it is self-related.
Accordingly, we have $\mathcal{N}_t' = \mathcal{N}_t \cup \left\{ r_{{\rm I}_{t,N_t +1}} \right\}$ and ${\mathcal{E}_t'} = {\mathcal{E}_t} \cup \left\{ 
{e_{{{\rm I}_{t,i}},{{\rm I}_{t,{N_t} + 1}}}}|1 \le i \le {N_t}+1
\right\}$ respectively.
We can obtain the updated adjacent and indegree matrix ${A_t'}=\left[a_{i,j} \right]_{1\le i,j\le N_t+1}$ and ${D_t'}= diag \left( {d_{{\rm I}_{t,1}}}, \cdots, {d_{{\rm I}_{t,i}}}, \cdots, {d_{{\rm I}_{t,N_{t}}}}, {d_{{\rm I}_{t,N_{t}+1}}}\right)$ respectively.
Hereafter, $\mathcal{G}_t'=(\mathcal{N}_t', \mathcal{E}_t')$ is a connected graph and $r_{{\rm I}_{t,N_t +1}}$ is capable of aggregating all local models and generates the global model $w_{t+1}$ after graph convolution.

% \begin{Ex2}[\textbf{ $D_t$ and $A_t$ }]
%     As is shown in Fig. \ref{aggregation}(a), we can obtain $D_t$ and $A_t$ as
%     \begin{equation}
%         {D_t} = \left[ {\begin{array}{*{20}{c}}
%             1&0&0&0&0\\
%             0&2&0&0&0\\
%             0&0&1&0&0\\
%             0&0&0&2&0\\
%             0&0&0&0&2
%         \end{array}} \right],
%         \notag
%     \end{equation}
%     \begin{equation}
%         {A_t} = \left[ {\begin{array}{*{20}{c}}
%             1&1&0&0&1\\
%             0&1&0&0&0\\
%             0&0&1&1&0\\
%             0&0&0&1&0\\
%             0&0&0&0&1
%         \end{array}} \right].
%         \notag
%     \end{equation}

% \end{Ex2}

% \begin{Ex3}[\textbf{ $D_t'$ and $A_t'$ }]
%     As is shown in Fig. \ref{aggregation}(b), we can obtain $D_t'$ and $A_t'$ as
%     \begin{equation}
%         {D_t'} = \left[ {\begin{array}{*{20}{c}}
%             1&0&0&0&0&0\\
%             0&2&0&0&0&0\\
%             0&0&1&0&0&0\\
%             0&0&0&2&0&0\\
%             0&0&0&0&2&0\\
%             0&0&0&0&0&6\\
%         \end{array}} \right],
%         \notag
%     \end{equation}
%     \begin{equation}
%         {A_t'} = \left[ {\begin{array}{*{20}{c}}
%             1&1&0&0&1&1\\
%             0&1&0&0&0&1\\
%             0&0&1&1&0&1\\
%             0&0&0&1&0&1\\
%             0&0&0&0&1&1\\
%             0&0&0&0&0&1\\
%         \end{array}} \right].
%         \notag
%     \end{equation}

% \end{Ex3}

\subsubsection{\textbf{Graph Convolution}}
Graph Convolution performs convolution directly on graphs\cite{kipfsemi}, which makes each node aggregate the information from its adjacent ones.
The conventional operation can be generally described as
\begin{equation}
{ Out} = \sigma ({ D}^{ - \frac{1}{2}} A{{ D}^{ - \frac{1}{2}}} L P),
\end{equation}
where $D$ and $A$ represent the indegree matrix and the adjacent matrix respectively. $L$ and $Out$ denote the input and the output of graph convolution. $\sigma(\cdot)$ represents the sigmoid function. $P$ denotes the parameter matrix which needs iterative training. 
In this mechanism, to prevent disturbance on the global model and simultaneously bring the benefits of reducing computational burden and execution time, we omit the sigmoid function as well as additional parameter matrix. 
Concretely, the graph convolution operation on $\mathcal{G}_t'$ can be formulated as
\begin{equation}
{\rm W_t'} =   {{ D_t'}^{ - \frac{1}{2}}} A_t' {{ D_t'}^{ - \frac{1}{2}}}{\rm W_t},
\end{equation}
where ${\rm W_t}$ and ${\rm W_t'}$ denote the set of raw and updated features. 

\textcolor{black}{For simplicity, we define the graph convolution operation as $M_t$. $M_t = { D_t'^{ - \frac{1}{2}}} A_t'{ D_t'^{ - \frac{1}{2}}}$ and $M_t=\left[m_{i,j} \right]_{1\le i,j\le N_t+1}$. }
The element ${m_{i,j}}$ represents the weight of $r_{{\rm I}_{t,i}}$ with respective to $r_{{\rm I}_{t,j}}$. 
According to the rules of matrix operations, we have 
\begin{align}
{m_{i,j}} = \left\{ {\begin{array}{*{20}{c}}
	{\frac{1}{{\sqrt {{d_{{{\rm I}_{t,i}}}}} }}\frac{1}{{\sqrt {{d_{{{\rm I}_{t,j}}}}} }},}&{{\rm if\textbf{\textbf{ }}}{ a_{i,j}} = 1,}\\
	{0,}&{\rm otherwise.}
	\end{array}} \right.
\end{align}

It is explicit that $m_{i,j}$ depends on ${d_{{\rm I}_{t,i}}}$, ${d_{{\rm I}_{t,j}}}$ and $a_{i,j}$.
Note that we only need to focus on the computation of $m_{i,N_t +1}$, due to the objective of generating the fresh global model $w_{t+1}$ at $r_{{\rm I}_{t,N_t +1}}$ after graph convolution.
With $a_{i,N_t +1} = 1$, the difference of $m_{i,N_t +1} (1\le i\le N_t+1)$ lies in the difference of ${{d_{{\rm I_{t,i}}}}}$.
${{d_{{{\rm I}_{t,i}}}}} > {{d_{{{\rm I}_{t,j}}}}} $ indicates that $r_{{\rm I}_{t,i}}$ has larger indegree, and therefore traffic data of $r_{{\rm I}_{t,i}}$ is influenced by more participants. 
Hence, $r_{{\rm I}_{t,i}}$ is less important in terms of spatial correlation compared with $r_{{\rm I}_{t,j}}$. 
It is reasonable that $m_{i,N_t +1} < m_{j,N_t +1}$. 
That is, $w_{t,{{\rm I}_{t,i}}}$ should be put a smaller weight than $w_{t,{{\rm I}_{t,j}}}$ when performing model aggregation.
When ${{d_{{{\rm I}_{t,i}}}}} = {{d_{{{\rm I}_{t,j}}}}} $, we deem $w_{t,{{\rm I}_{t,i}}}$ and $w_{t,{{\rm I}_{t,j}}}$ have equal importance in aggregation. 

However, the above strategy is unreasonable since the outdegrees of $r_{{\rm I}_{t,i}}$ and $r_{{\rm I}_{t,j}}$ are not considered. The outdegree of a certain participant can evaluate how many participants it affects.
Therefore, participants with larger outdegree should be paid more attention in model aggregation.
In fact, this problem is subject to the inherent characteristic of graph convolution, where receptive fields depend on graph convolution layers.
That is, after $g$-layer graph convolution, each participant can aggregate features from the ones whose distance to it is $g$ steps. 
%For $r_{{\rm I}_{t,N_t+1}}$, each participant is adjacent to it, and hence it is capable of aggregating all local updated models after 1-layer graph convolution.
Since all participants are adjacent to $r_{{\rm I}_{t,N_t+1}}$, $r_{{\rm I}_{t,N_t+1}}$ is capable of aggregating all local updated models after 1-layer graph convolution.
However, $r_{{\rm I}_{t,N_t+1}}$ fails to capture the adjacency among real participants, as is shown in Fig. \ref{aggregation} (c).
Therefore, we propose to leverage 2-layer graph convolution for $r_{{\rm I}_{t,N_t+1}}$ to obtain the adjacency among all participants. 

%Naturally, we consider to increase the layers of graph convolution.
\textcolor{black}{We define $V_t =\left({ D_t'^{ - \frac{1}{2}}} A_t'{ D_t'^{ - \frac{1}{2}}}\right)^2$ as two-layer graph convolution, and $V_t=\left[v_{i,j} \right]_{1\le i,j\le N_t+1}$}. 
Likewise, we only focus on the values of \textcolor{black}{$v_{i,N_t+1} \left({1\le i\le N_t+1}\right)$} and have 
\textcolor{black}{
\begin{equation}
v_{i,{N_t} + 1} = \sum\limits_{k = 1}^{{N_t} + 1} {{m_{i,k}}{m_{k,{N_t} + 1}}}.
\end{equation}
}

It is intuitive that the calculation of \textcolor{black}{$v_{i,{N_t} + 1}$} involves all 2-step paths from $r_{{\rm I}_{t,i}}$ to $r_{{\rm I}_{t,N_t+1}}$.
According to graph construction, the longest path from $r_{{\rm I}_{t,i}} (1\le i\le N_t+1)$ to $r_{{\rm I}_{t,N_t+1}}$ is 2 steps.
Therefore, 2-layer graph convolution is enough to obtain the inner spatial correlation among multiple participants.
%Therefore, 2-layer graph convolution is enough for $r_{{\rm I}_{t,N_t+1}}$ to obtain the adjacency among all participants. 
Besides, the execution of graph convolution excludes parameters to be trained iteratively and is just based on matrix multiplication operation on the server side, which is regarded as computation-lightweight.
In addition, this method doesn't need participants to transmit extra parameters, thus evidently decreasing communication overhead.

\subsubsection{\textbf{Model Aggregation}}

We first perform normalization as
\textcolor{black}{
\begin{equation}
v_{i,{N_t} + 1} = v_{i,{N_t} + 1} / sum_t, \left({1\le i\le N_t+1}\right),
\end{equation}
where $su{m_t} = \sum\nolimits_{i = 1}^{{N_t} + 1} {v_{i,{N_t} + 1}} $.
We deem that $v_{i,{N_t} + 1}$ can efficiently evaluate the importance of local models trained on $r_{\rm I_{t,i}}$ for model aggregation, and thus is directly regarded as the aggregation weight of $w_{t,\rm I_{t,i}}$. Therefore, the fresh global model $w_{t+1}$ is ultimately yielded as
\begin{equation}
{w_{t+1}} = {w_{t}}v_{{N_t} + 1,{N_t} + 1} + \sum\limits_{i = 1}^{{N_t}} {{w_{{t},{{\rm I}_{t,i}}}}v_{i,{N_t} + 1}} .
\end{equation}
}
\subsection{Execution Process of REFOL}
In this subsection, the whole procedure of REFOL is elaborated in Algorithm \ref{ag3}. 
At the beginning of each round, each client firstly detects concept drift by evaluating distribution difference of traffic data between the current and previous rounds using KLD, followed by comparing with threshold $Q$ to determine whether to download the fresh global model or reuse the locally-saved model for online forecasting using Algorithm \ref{ag2}. 
After then the participants upload the updated models to the aggregator for aggregation.
The aggregator calculates the aggregation weights of participants via 2-layer graph convolution, and then performs model aggregation to generate the fresh global model (Line 5-8).

\begin{algorithm}
	\caption{REFOL}
	\label{ag3}
	\LinesNumbered
	\KwIn{Initialized model parameter $w_1$, learning rate $\eta$,  threshold $Q$, client set $\mathcal{N}$.}
	\KwOut{The global model $w_{\mathcal{T}+1}$.}
	
	\SetKwFunction{Fexecute}{ClientExecute}
	
	\textsc{\textbf{ServerExecute:}}\\
	\For {$t = 1, 2, \cdots, \mathcal{T}$}{
		\For {$r_n \in \mathcal{N}$ in parallel}{
			\Fexecute($n$, $t$)\\
		}
		Receive local updated models from the participants.\\
		\textcolor{black}{Perform 2-layer graph convolution via Eq. (13)(14).}\\
		\textcolor{black}{Normalize aggregation weights via Eq. (15).}\\
		\textcolor{black}{Perform aggregation via Eq. (16) and generate $w_{t+1}$.}\\
	}
    \KwRet $w_{\mathcal{T}+1}$
	
	\SetKwProg{Fn}{Function}{:}{}
	\Fn{\Fexecute{$n$, $t$}}{
		\textcolor{black}{Execute local processes as Algorithm \ref{ag2}}.
	}
\end{algorithm}

\section{Experiments}
In this section, comprehensive experiments are conducted to validate the high efficiency of our proposed REFOL. Firstly, we present a brief introduction of the system configurations including datasets, metrics, experiment settings, and baselines. Then we analyze the performance comparisons of REFOL and baselines. Finally, the ablation study is conducted and the effects of varying parameter settings on prediction performance are further explored.

\subsection{Experimental Settings}
\subsubsection{\textbf{Datasets and Metrics}}
% todo
The experiment datasets are generated from two real-world datasets: PEMS-BAY and METR-LA \cite{li2018diffusion} \footnote{\textcolor{black}{The datasets can be downloaded \href{https://drive.google.com/drive/folders/10FOTa6HXPqX8Pf5WRoRwcFnW9BrNZEIX}{here}}}. PEMS-BAY includes the vehicular speed from 325 sensors in the Bay Area from 2017/1/1 to 2017/5/31. METR-LA is comprised of traffic speed collected by 207 sensors in the highway of Los Angeles County from 2012/3/1 to 2012/6/30.
% For the two datasets, we select the traffic data from 50 sensors on Sundays to Thursdays in May as experiment datasets. 
The adjacent matrices of sensors are constructed as per \cite{li2018diffusion}. Two widely-used metrics for regression tasks, Root Mean Square Error (RMSE) and Mean Absolute Error (MAE) are adopted to assess the prediction performance. RMSE and MAE are calculated respectively as follows:
\begin{equation}
    {\rm{RMSE}} = \frac{{\sum\nolimits_{n = 1}^N {\sum\nolimits_{t = 1}^{\mathcal{T}} {\sqrt {\frac{{\sum\nolimits_{\tau  = 1}^F {{{({s_{t + \tau ,n}} - {{\hat s}_{t + \tau ,n}})}^2}} }}{F}} } } }}{{N{\mathcal{T}}}},
\end{equation}
\begin{equation}
    {\rm{MAE}} = \frac{{\sum\nolimits_{n = 1}^N {\sum\nolimits_{t = 1}^{\mathcal{T}} {\frac{{\sum\nolimits_{\tau  = 1}^F {\left| {{s_{t + \tau ,n}} - {{\hat s}_{t + \tau ,n}}} \right|} }}{F}} } }}{{N{\mathcal{T}}}}.
\end{equation}

\subsubsection{\textbf{System Configuration}}
All experiments are conducted on a server with Intel(R) Xeon(R) Gold 5218 CPU @ 2.30GHz and two NVIDIA Geforce RTX 3090 Founders Edition GPUs. We implement REFOL with PyTorch, where a GRU network with 1 layer with 128 hidden cells is used as the prediction model. 
The learning rate is set to 0.001 and the local epoch $E$ is set to 5. The KLD threshold $Q$ is set to $0.0003$ (how to determine such threshold is analyzed in Section \ref{Q}). \textcolor{black}{The historical horizon $H$ is set to 12.
We adopt MSE as the loss function.}
We divide the traffic flow series into samples by sliding window strategy, as per \cite{meng2021cross}.
For batch learning methods, all samples are split to train, validation, and test datasets with the ratio of 7:1:2. For online learning, we average the prediction errors of the samples belonging to the test dataset in the batch learning methods to achieve fair comparison.
% todo
\textcolor{black}{
The source code is available at \href{https://github.com/yuppielqx/REFOL}{https://github.com/yuppielqx/REFOL}.
}

\subsubsection{\textbf{Baselines}}
\textcolor{black}{We compare the prediction performance of REFOL with that of 12 baselines, with 5 centralized methods (SVR\cite{feng2006svm}, GRU\cite{dai2019short}, DCRNN\cite{li2018diffusion}, STGCN\cite{yu2018spatio}, and MegaCRN\cite{jiang2023spatio}) and 7 FL methods (FedAvg\cite{mcmahan2017communication}, FedGRU\cite{liu2020privacy}, FCGCN\cite{xia2022short}, FASTGNN\cite{zhang2021fastgnn}, CNFGNN\cite{meng2021cross}, FedGTP\cite{yang2024fedgtp}, and pFedCTP\cite{zhang2024personalized}).}
Note that all of these baselines are trained in offline learning and the configurations of baselines comply with the corresponding literature, unless stated otherwise.
The brief introduction of these baselines is presented as follows.

\begin{itemize}
    \item SVR (Support Vector Regression) \cite{feng2006svm}: 
    %SVR is a classical machine learning algorithm. In our experiments,
    Each client independently predicts traffic flow using SVR and doesn't interact with the central server.
    
	\item {GRU}\cite{dai2019short}: The central server conducts prediction using GRU based on the uploaded raw traffic data.
    \textcolor{black}{
    \item {DCRNN}\cite{li2018diffusion}: It is a centralized prediction method, which treats traffic flow as a diffusion process and also evaluates spatio-temporal correlation among clients.}
    \textcolor{black}{
    \item STGCN \cite{yu2018spatio}: It is centralized method, where the graph convolution and gated temporal convolution are adopted for fast training and accurate performance.}
    \textcolor{black}{
    \item MegaCRN \cite{jiang2023spatio}: It is a centralized method, where meta-graph convolutional recurrent network is proposed to tackle the spatio-temporal heterogeneity.}
    \textcolor{black}{
    \item FedAvg \cite{mcmahan2017communication}: It is a conventional framework, where the central server aggregates the locally-updated model parameters via averaging mechanism.}

    \item{
        {FedGRU} \cite{liu2020privacy}: Traffic nodes are divided into clusters and train local GRU models. The server conducts aggregation by averaging mechanism.
    }
    \item{
        {FCGCN} \cite{xia2022short}: Traffic nodes are divided into local road networks, each of which is treated as a client in the FL paradigm and trains a GCN model. The server aggregates the GCN model parameters by averaging mechanism.
    }
    \item{
        {FASTGNN} \cite{zhang2021fastgnn}: Like FedGRU, each cluster works as FL client which adopts GAT to assess the spatial correlation and GRU for prediction. The server aggregates the adjacency matrix and model parameters across clusters.
    }
	\item {CNFGNN}\cite{meng2021cross}: 
	It is a FL method, which adopts encoder-decoder architecture of GRU and GNN to predict traffic flow based on evaluating spatio-temporal correlation.  
    \textcolor{black}{
    \item FedGTP \cite{yang2024fedgtp}: It is a novel federated graph-based traffic prediction framework for better capturing intra-client spatial dependencies.}
    \textcolor{black}{
    \item pFedCTP \cite{zhang2024personalized}: It is a personalized federated learning method for cross-city traffic prediction, which aims to construct personalized prediction models for data-sparse cities by knowledge transfer.}
\end{itemize}

\subsection{Main Results}

\begin{table*}[!t]
  \centering
  \caption{\textcolor{black}{Comparison in Prediction Performance on Two Datasets with Different Forecasting Steps}}
  \renewcommand\arraystretch{1.4}
%   \small
    \begin{tabular}{c||cc|cc|cc||cc|cc|cc}
    \hline
    \multirow{2}[6]{*}{Methods} & \multicolumn{6}{c||}{PEMS-BAY} &  \multicolumn{6}{c}{METR-LA} \\ 
    \cline{2-13}  & \multicolumn{2}{c|}{5min ($F=1$)} & \multicolumn{2}{c|}{30min ($F=6$)} & \multicolumn{2}{c||}{1h ($F=12$)} & \multicolumn{2}{c|}{5min ($F=1$)} & \multicolumn{2}{c|}{30min ($F=6$)} & \multicolumn{2}{c}{1h ($F=12$)}  \\
\cline{2-13} & RMSE  & MAE & RMSE  & MAE & RMSE  & MAE & RMSE  & MAE & RMSE  & MAE & RMSE  & MAE \\
    \hline
    SVR & 2.01  & 1.09  & 3.74  & 1.79  & 4.80  & 2.28 & 6.43  & 3.46  & 8.41  & 4.34  & 9.61  & 5.00 \\
    GRU & 2.02  & {0.99}  & 3.93  & 1.77  & {5.50}  & {2.42} & 5.56  & \textbf{\textcolor{black}{2.91}}  & 8.13  & {4.17}  & 9.41  & {4.90} \\
    \textcolor{black}{DCRNN} & \textcolor{black}{1.65}  & \textcolor{black}{0.93}  & \textcolor{black}{4.79}  & \textcolor{black}{2.09}  & \textcolor{black}{6.19}  & \textcolor{black}{2.71} & \textcolor{black}{4.30}  & \textcolor{black}{3.57}  & \textcolor{black}{7.09}  & \textcolor{black}{4.61}  & \textcolor{black}{8.71}  & \textcolor{black}{5.33} \\
    \textcolor{black}{STGCN} & \textcolor{black}{1.46}  & \textcolor{black}{\textbf{0.86} } & \textcolor{black}{3.62}  & \textcolor{black}{1.73}  & \textcolor{black}{4.43}  & \textcolor{black}{2.26} & \textcolor{black}{5.92}  & \textcolor{black}{3.82}  & \textcolor{black}{8.60}  & \textcolor{black}{4.62}  & \textcolor{black}{10.66}  & \textcolor{black}{5.92}  \\
    \textcolor{black}{MegaCRN} & \textcolor{black}{1.61}  & \textcolor{black}{0.91}  & \textcolor{black}{4.42}  & \textcolor{black}{1.94}  & \textcolor{black}{5.33}  & \textcolor{black}{2.33} & \textcolor{black}{4.24}  & \textcolor{black}{3.54}  & \textcolor{black}{6.77}  & \textcolor{black}{4.51}  & \textcolor{black}{8.11}  & \textcolor{black}{5.09} \\
    \hline
    \textcolor{black}{FedAvg} & \textcolor{black}{1.80}  & \textcolor{black}{1.02}  & \textcolor{black}{3.72}  & \textcolor{black}{1.78}  & \textcolor{black}{5.25}  & \textcolor{black}{2.41} & \textcolor{black}{5.84}  & \textcolor{black}{3.32}  & \textcolor{black}{8.16}  & \textcolor{black}{4.45}  & \textcolor{black}{9.52}  & \textcolor{black}{5.22} \\
     {FedGRU} &  {1.82} &  0.97 &  {4.72} &  {2.27} &  {6.52} &  {3.37} & {5.71} &  {2.99} &  {9.23} &  {5.00} &  {11.03} &  {6.56} \\
     {FCGCN} &  {9.25} &	 {4.82} &  {9.45} &	 {4.97} &  {9.36} &	 {5.04} &  {15.37} &  {9.69} &  {15.45} &  {10.04} &  {15.16} &  {9.83} \\
     {FASTGNN} &  {5.32} &  {2.82} &  {6.06} &  {3.20} &  {6.84} &  {3.76} &  {11.73} &  {7.27} &  {12.98} &  {8.23} &  {12.82} &  {8.16} \\
    CNFGNN & {1.82}  & {1.04}  & {3.60}  & {1.73}  & 5.13  & 2.64 & {5.82}  & {3.25}  & {8.03}  & 4.24  & {9.38}  & 5.12 \\
    \textcolor{black}{FedGTP} & \textcolor{black}{1.75}  & \textcolor{black}{1.00} & \textcolor{black}{4.59}  & \textcolor{black}{2.06}  & \textcolor{black}{6.10}  & \textcolor{black}{2.74} & \textcolor{black}{6.11}  & \textcolor{black}{2.96 } & \textcolor{black}{9.34}  & \textcolor{black}{4.40}  & \textcolor{black}{11.77}  & \textcolor{black}{5.66} \\
    \textcolor{black}{pFedCTP} & \textcolor{black}{5.29}  & \textcolor{black}{2.84}  & \textcolor{black}{6.49}  & \textcolor{black}{3.40}  & \textcolor{black}{7.43}  & \textcolor{black}{3.92} & \textcolor{black}{8.88}  & \textcolor{black}{5.21}  & \textcolor{black}{10.56}  & \textcolor{black}{6.02}  & \textcolor{black}{12.08}  & \textcolor{black}{7.15} \\
    \hline 
    {REFOL} & \textbf{1.00}  & {1.00}  & \textbf{{1.86}}  & \textbf{{1.61}}  & \textbf{{2.44}}  & \textbf{{2.08}} & \textbf{{3.29}}  & 3.29 & \textbf{{4.87}}  & \textbf{{4.02}}  & \textbf{{5.29}}  & \textbf{{4.21}}  \\
    \hline
    \end{tabular}%
  \label{result}%
\end{table*}%

\begin{figure*}[!ht]
	\centering
	\includegraphics[width=1\textwidth]{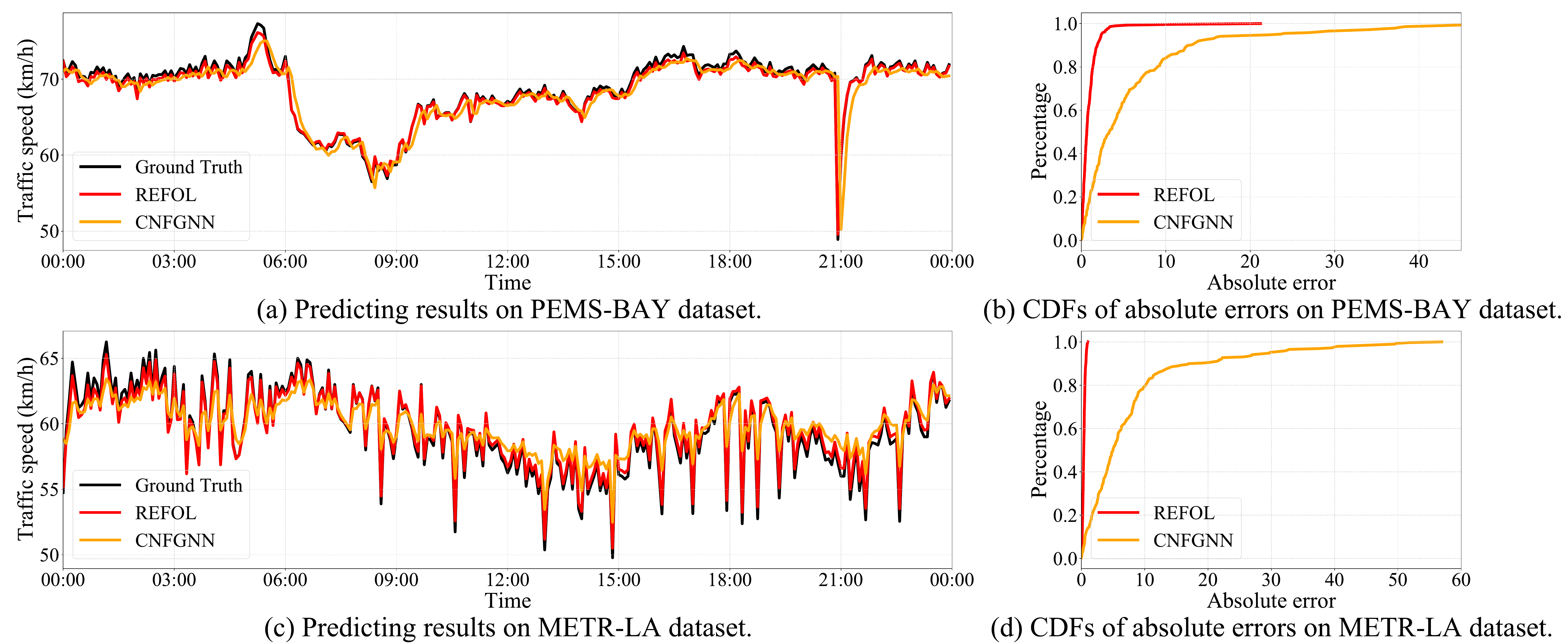}
	\caption{\textcolor{black}{Ground truth values and forecasting values of CNFGNN and REFOL.}}
	\label{y_and_y_pred}
\end{figure*}

The RMSE and MAE results of different methods are summarized in Table \ref{result}. We compare the predicting errors of the \textcolor{black}{12} methods with different forecasting steps, i.e., $F=1$, $F=6$ and $F=12$. 
RMSE and MAE values of REFOL on two datasets are equal in the case of $F=1$. It is because that when $F=1$, we have 
\begin{equation}
    {\sqrt {\frac{{\sum\nolimits_{\tau  = 1}^F {{{({s_{t + \tau ,n}} - {{\hat s}_{t + \tau ,n}})}^2}} }}{F}} }
    = \left| s_{t+1,n} - \hat{s}_{t+1,n}   \right|.
\end{equation}

We have the key observations from Table \ref{result} that the proposed REFOL performs best among the baselines on both datasets. 
%怎样提升，提升了多少
Specifically, compared with state-of-the-art FL method CNFGNN, REFOL can offer RMSE gains of 45.05\%, 48.33\%, and 52.44\% on PEMS-BAY dataset when $F=1$, $F=6$, and $F=12$ respectively. 
Likewise, for METR-LA dataset, REFOL obtains RMSE gains of 43.43\% ($F=1$), 39.35\% ($F=6$), and 43.60\% ($F=12$) respectively.

We compare predicted values of different methods. The predicted results and cumulative distribution functions (CDFs) of absolute errors are shown in Fig. \ref{y_and_y_pred}. 
% \textcolor{black}{In Fig. \ref{y_and_y_pred} (a) and (c), the prediction and ground truth values are de-normalized for explicitly show the difference in forecasting values.}
\textcolor{black}{As is illustrated in Fig. \ref{y_and_y_pred}(a) and (c), the predicted values of REFOL have the same distribution with the ground truth on both datasets.}
In terms of prediction errors, REFOL yields much smaller absolute errors, compared with CNFGNN.
\textcolor{black}{As shown in Fig. \ref{y_and_y_pred} (b), on PEMS-BAY dataset, 67\% of the absolute errors in REFOL are smaller than 1, while 20\% in CNFGNN. 
As shown in Fig. \ref{y_and_y_pred} (d), the comparisons on METR-LA dataset are more remarkable, with 100\% and 14\% for REFOL and CNFGNN respectively.}
Hence, we can conclude that REFOL yields superior prediction performance, compared with state-of-the-art FL method and the other offline methods.

\begin{figure*}[!ht]
	\centering
	\includegraphics[width=\textwidth]{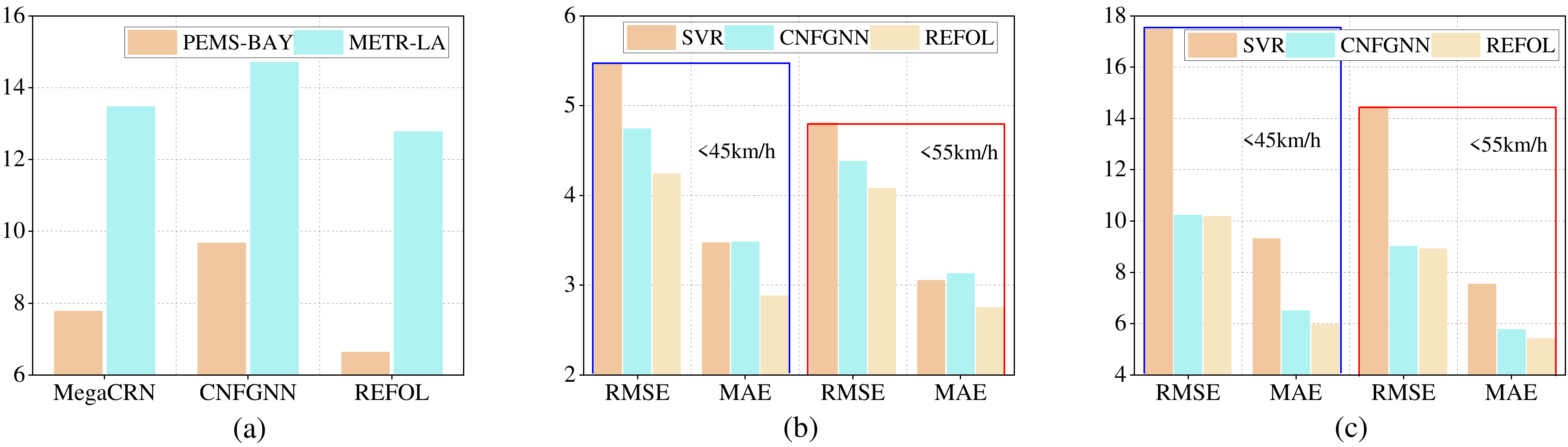}
	\caption{\textcolor{black}{
    (a) Further performance comparison in prediction intervals; (b) and (c): Prediction performance in the conditions of traffic jams on PEMS-BAY and METR-LA datasets.}}
	\label{error_bar}
\end{figure*}

\begin{figure*}[!ht]
	\centering
	\includegraphics[width=0.9\textwidth]{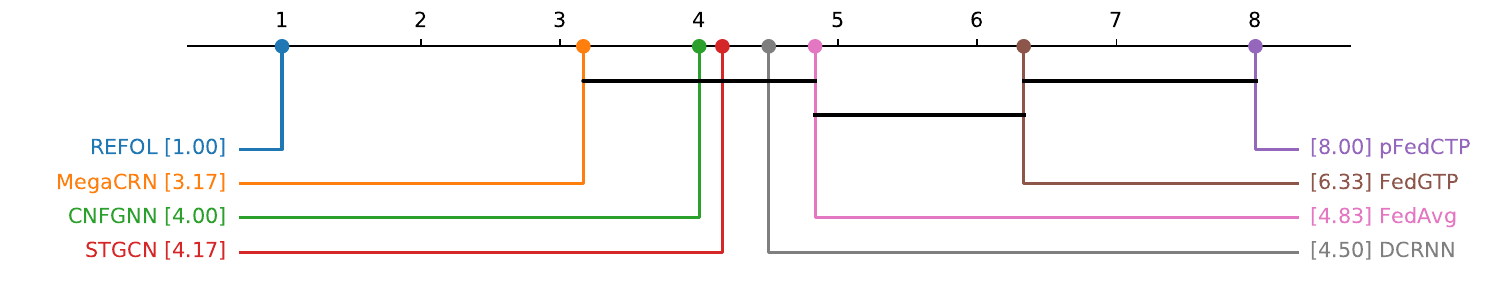}
	\caption{\textcolor{black}{The mean ranks of methods with different forecasting steps on two datasets.
    A horizontal bold line indicates that there is no significant difference in prediction performance among the corresponding methods.}}
	\label{CD}
\end{figure*}

\textcolor{black}{
Moreover, we employ mean scaled interval score (MSIS) to evaluate the prediction intervals of each method \cite{gneiting2007strictly,SPILIOTIS2020550}, which is formulated as:
\begin{align}
    scale &= \frac{1}{{(\mathcal{T} - \mathcal{P})N}}\sum\limits_{n = 1}^N {\sum\limits_{t = \mathcal{P} + 1}^\mathcal{T} {|{s_{t,n}} - {s_{t - \mathcal{P},n}}|} }, \\
    width &= \frac{1}{{N\mathcal{T}}}\sum\limits_{n = 1}^N {\sum\limits_{t = 1}^\mathcal{T} {(\hat s_{t,n}^u - \hat s_{t,n}^l)} }, \\
    p_l &= \frac{1}{{N\mathcal{T}}}\sum\limits_{n = 1}^N {\sum\limits_{t = 1}^\mathcal{T} {(\hat s_{t,n}^l - {s_{t,n}})\boldsymbol{1}\{ {s_{t,n}} < \hat s_{t,n}^l\} } }, \\
    p_u &= \frac{1}{{N\mathcal{T}}}\sum\limits_{n = 1}^N {\sum\limits_{t = 1}^\mathcal{T} {({s_{t,n}} - \hat s_{t,n}^u)\boldsymbol{1}\{ {s_{t,n}} > \hat s_{t,n}^u\} } }, \\
    {\rm MSIS} &= \frac{1}{{scale}}(width + \frac{2}{\alpha }({p_l} + {p_u})),
\end{align}
where $\hat{s}_{t,n}^l$ and $\hat{s}_{t,n}^u$ denote the lower and upper prediction bounds of $s_{t,n}$. $\mathcal{P}$ represents the periodicity of the traffic flow. $\alpha$ denotes the significance level. $\boldsymbol{1}\{\cdot\}$ is the indicator function (being 1 if $\cdot$ is true and 0 otherwise).
}
\textcolor{black}{
We compare the MSIS results of REFOL with those of the top centralized method MegaCRN and the top FL method CNFGNN, which is shown in Fig. \ref{error_bar} (a).
We have the key observation that REFOL provides the lowest MSIS results on both datasets, i.e., narrower prediction intervals, which demonstrates that REFOL can promise more precise prediction.
}

\textcolor{black}{
Given that traffic control centers may be more interested in accurate prediction during congested time periods, we investigate the prediction performance in traffic congestion.
When the traffic speed is lower than 45 km/h or 55 km /h, we deem that traffic congestion occurs. We respectively calculate the prediction errors when true traffic speed values do not exceed such criteria.
As shown in Fig. \ref{error_bar} (c), REFOL can always achieve the superior performance, especially on PEMS-BAY dataset, with 10.5\% (4.6\%) RMSE reduction compared with CNFGNN when 45 km/h (55 km/h) is adopted as the criterion.
The performance gains result from the effectiveness of  evaluating spatio-temporal correlation in a fine-grained way. 
}

% todo
\textcolor{black}{
We further incorporate the statistical significance testing to evaluate the performance significance. The critical difference diagram in Fig. \ref{CD} shows the averaged ranks of each model's prediction RMSEs with different forecasting step on two datasets. We have the key observation that REFOL has the lowest rank, i.e., the best prediction performance. Moreover, REFOL outperforms the baselines significantly (no horizontal bold lines across REFOL and any other methods), which validates 
REFOL can promise impressive performance gains.
}

\subsection{Comparison of Computational and Communication Cost at Clients}
\begin{table*}[!htbp]
  \centering
  \caption{Comparisons in Computational and Communication Cost at Clients}
  \renewcommand\arraystretch{1.4}
  \small
    \begin{tabular}{c||cc|cc|cc|c|c}
    \hline
    \multirow{3}[6]{*}[+1.5ex]{Methods} & \multicolumn{8}{c}{PEMS-BAY} \\
\cline{2-9}          & \multicolumn{2}{c|}{5min (F=1)} & \multicolumn{2}{c|}{30min (F=6)} & \multicolumn{2}{c|}{1h (F=12)} & \multirow{2}[4]{*}[+1.5ex]{\makecell[c]{Computational  Cost/(GFLOPs)}} & \multirow{2}[4]{*}[+1.5ex]{\makecell[c]{Communication  Cost/(GB)}} \\
\cline{2-7}          & RMSE  & MAE   & RMSE  & MAE   & RMSE  & MAE   &       &  \\
    \hline
    FedOSTC & 0.96  & 0.96  & {1.74}  & {1.51}  & {1.91}  & {1.62}  & 533.45 & 303.41 \\
    FOL-vanilla & 6.55  & 6.55  & 6.73  & 6.47  & 6.87  & 6.41  & {125.00}   & {37.72} \\
    % FedOnline & 0.98  & 0.98  & 1.46  & 1.18  & 1.51  & 1.15  & 361.11 & 134.36 \\
    REFOL & {1.00}     & {1.00}     & 1.86  & 1.61  & 2.44  & 2.08  & 125.03 & {37.72} \\
    \hline
    \hline
    \multirow{3}[6]{*}[+1.5ex]{Methods} & \multicolumn{8}{c}{METR-LA} \\
\cline{2-9}          & \multicolumn{2}{c|}{5min (F=1)} & \multicolumn{2}{c|}{30min (F=6)} & \multicolumn{2}{c|}{1h (F=12)} & \multirow{2}[4]{*}[+1.5ex]{\makecell[c]{Computational  Cost/(GFLOPs)}} & \multirow{2}[4]{*}[+1.5ex]{\makecell[c]{Communication  Cost/(GB)}} \\
\cline{2-7}          & RMSE  & MAE   & RMSE  & MAE   & RMSE  & MAE   &       &  \\
    \hline
    FedOSTC & {2.90}  & {2.90}  & {4.35}  & {3.66}  & {4.98}   & {4.10}  & 533.45 & 303.41 \\
    FOL-vanilla & 11.72 & 11.72 & 12.22 & 11.69 & 12.55 & 11.81 & {310.65} & {113.71} \\
    % FedOnline & 3.18  & 3.18  & 4.31  & 3.33  & 4.45  & 3.23  & 361.11 & 134.36 \\
    REFOL & 3.29  & 3.29  & 4.87  & 4.02  & 5.29  & 4.21  & 310.68 & {113.71} \\
    \hline
    \end{tabular}%
  \label{online_comparison}%
\end{table*}%

We compare the communication and computational cost of all clients in REFOL with the other two FOL prediction methods. These FOL methods are introduced as follows.
\begin{itemize}
    \item FedOSTC \cite{10618965}: All clients are selected as participants at each round. Each client trains an encoder-decoder architecture of GRU locally. The server maintains a GAT for evaluating the spatial correlation and aggregates local models via averaging strategy.

    \item FOL-vanilla: The server randomly selects a certain number of clients as participants at each round. For fair comparison, the participant number is equal to the averaged participant number over $\mathcal{T}$ rounds in REFOL.

    % \item FedOnline: This method is the same with REFOL, except that all clients are selected as participants in each round. 
\end{itemize}

\subsubsection{\textbf{Computational Cost Analysis}}
Adaptive online optimization contains two functions, i.e., prediction and local optimization. 
In fact, these two functions can be fulfilled by the processes of $forward$ $propagation$ and $backward$ $propagation$ in the context of machine learning. 
Hence, we calculate the computational cost of the two processes.
The forward propagation of GRU is shown in Eq. (6)-(9) and we ignore the computational cost generated by Eq. (7).
According to \cite{molchanov2019pruning}, the computational cost of Eq. (6)-(8) can be calculated as $\left( {1 + hs} \right) \times hs \times 3 \times 2 ({\rm FLOPs})$.
The computational cost of the fully connected layer is $ hs \times 2 ({\rm FLOPs})$. The computational cost of one forward propagation in GRU is 
\begin{equation}
    \left( {1 + hs} \right) \times hs \times 3 \times 2 + hs  \times 2 ({\rm FLOPs}).
\end{equation}
The computational cost of backward propagation is twice as much as that of the forward propagation \cite{epoch2021backwardforwardFLOPratio}. Hence, the computational cost of one backward propagation is 
\begin{equation}
    2 \times \left(\left( {1 + hs} \right) \times hs \times 3 \times 2 + hs  \times 2 \right) ({\rm FLOPs}).
\end{equation}
In FedOSTC, the encoder and decoder are both GRUs. Hence, we can calculate the computational cost accordingly as above.

Moreover, we consider the computational cost in calculating KLD values.
% According to Eq. (6)(7), t
The process can be split into three operations as 
\begin{itemize}
    \item $r_n$ calculates the summation of $s_{t-i,n}(0\le i\le H-1)$ and $s_{u-i,n}(0\le i\le H-1)$, and the corresponding computational cost is $H\times 2{\rm (FLOPs)}$.
    \item $r_n$ divides $s_{t-i,n}(0\le i\le H-1)$ and $s_{u-i,n}(0\le i\le H-1)$ by ${{\sum\nolimits_{i = 0}^{H - 1} {{s_{t - i,n}}} }}$ and ${{\sum\nolimits_{i = 0}^{H - 1} {{s_{u - i,n}}} }}$ respectively. The computational cost is $H\times 2{\rm (FLOPs)}$.
    \item $r_n$ finally calculates the KLD value, which is composed by operations of division, logarithm, multiplication, and addition. The generated computational cost is $H\times 3{\rm (FLOPs)}$.
\end{itemize}

Therefore, the computational cost of calculating KLD value is $H\times 7{\rm (FLOPs)}$.

\subsubsection{\textbf{Communication Cost Analysis}}
Then we dive into how to calculate the communication cost of all clients in different methods.
For FOL-vanilla and REFOL, the participants just need to transmit model parameters of the prediction models (i.e., GRU network).
Based on Eq. (6)-(9), the model parameter amount of GRU layer can be calculated as$(3 \times hs + 3 \times hs \times hs + 3 \times hs)$.
In addition, the parameter amount of the fully connected layer is $(hs + 1)$. Therefore, the parameter amount of prediction model can be calculated as
\begin{equation}
    3 \times hs \times (hs + 2) + hs + 1.
\end{equation}
Since the encoder and decoder module in FedOSTC are all GRU networks, the model parameter amount can also be calculated accordingly.
Furthermore, the additional parameters should be exchanged between clients with the aggregator in FedOSTC, for updating clients' hidden states and optimizing GNN parameters, which can be calculated as follows \cite{meng2021cross}:
\begin{equation}
    hs + [3 \times 2hs \times (2hs + 2) + (2hs + 1) ].
\end{equation}

\subsubsection{\textbf{Comparison Analysis}}
The experimental results on prediction performance, computational cost, and communication cost at clients are presented in Table \ref{online_comparison}.
It is illustrated that the FOL-vanilla performs worst in terms of perdiction accuracy, in spite of the equal participants number as REFOL.
\textcolor{black}{In FedOSTC, all clients participate in each federated training round. Therefore, the computational and communication costs keep constant on two datasets. While in REFOL, each client need to determine whether to participate in federated round based on traffic distribution, the cost reductions on different datasets vary, resulting variant computational and communication costs on PEMS-BAY and METR-LA datasets.}
 
Although FedOSTC achieves the best prediction accuracy, compared with FOL-vanilla and REFOL, the yielded computational cost and communication cost are much higher than the other two methods, mainly because all clients are selected as participants at each training round and extra parameters should be transmitted for updating hidden states and for optimizing GAT's parameters.
Compared with FOL-vanilla, REFOL has an extra lightweight process of calculating KLD values at clients. Therefore, the total computational cost of clients in REFOL is little higher than that in FOL-vanilla.
But more importantly, REFOL can guarantee prediction performance, with a small drop from the best-performed FedOSTC, while significantly decreasing the computational and communication costs by {$\mathbf{76.56\%}$} and $\mathbf{87.57\%}$ ($\mathbf{41.76\%}$ and $\mathbf{62.52\%}$) on PEMS-BAY dataset (METR-LA dataset) respectively.
This improvement results from the designed data-driven client participation mechanism to avoid redundant model updates and graph convolution-based model aggregation to integrate the process of evaluating spatial correlation into model aggregation.

\subsection{Ablation Study}
\begin{table*}[!ht]
  \centering
  \caption{Ablation Tests on PEMS-BAY Dataset}
  \renewcommand\arraystretch{1.4}
  \small
    \begin{tabular}{ccc||cc|cc|cc}
    \hline
     \multirow{2}[4]{*}[+1.5ex]{Methods} & \multirow{2}[4]{*}[+1.5ex]{\makecell[c]{Participation  determination}} & \multirow{2}[4]{*}[+1.5ex]{\makecell[c]{Aggregation strategy}} & \multicolumn{2}{c|}{5min (F=1)} & \multicolumn{2}{c|}{30min (F=6)} & \multicolumn{2}{c}{1h (F=12)} \\
\cline{4-9}          &       &       & RMSE  & MAE   & RMSE  & MAE   & RMSE  & MAE \\
    \hline
    REFOL-D & \emph{data-driven} & \ding{55}    & 1.22  & 1.22  & 2.53  & 2.26  & 2.99  & 2.60 \\
    % REFOL-V1 & \emph{random selecion-based} &  \emph{averaging} & 6.55  & 6.55  & 6.73  & 6.47  & 6.87  & 6.41 \\
    REFOL-V1 & \emph{random selection-based} & \emph{graph convolution-based} & 6.62  & 6.62  & 6.72  & 6.46  & 6.93  & 6.48 \\
    REFOL-V2 & \emph{data-driven} &  \emph{averaging} & {1.00}     & {1.00}     & 2.10   & 1.82  & 2.73  & 2.32 \\
    REFOL & \emph{data-driven} & \emph{graph convolution-based} & {1.00}     & {1.00}     & {1.86}  & {1.61}  & {2.44}  & {2.08} \\
    \hline
    \end{tabular}%
  \label{ablation}%
\end{table*}%

In this subsection, we conduct ablation study to verify the effectiveness of different components in REFOL.
Firstly, we transform the REFOL method into the following variants.
\begin{itemize}
    \item REFOL-D: Each client online performs prediction distributedly. When concept drift occurs, clients independently optimize their locally-owned models.
    \item REFOL-V1: The central server randomly selects participants at each round and aggregates local models based on our designed aggregation mechanism.
    We calculate the average number of participants per round in REFOL as the selected number in REFOL-V1.
    \item REFOL-V2: It is the same with our designed REFOL except that the server aggregates local models based on the averaging strategy.
\end{itemize}
We conduct extensive experiments on PEMS-BAY dataset to compare the prediction performance of different variants and the predicting results are summarized in Table \ref{ablation}.

It is plain that REFOL performs best among all the variants with different forecasting steps. 
\textcolor{black}{Compared with the three variants, the averaged (RMSE, MAE) gains of REFOL are (1.95, 1.95), (1.92, 1.90), and (1.78, 1.72)  with $F=1, 6,$ and 12 respectively.
Therefore, the smaller the forecasting step is, the more performance gains REFOL will offer.
}

Specifically, compared with REFOL-D, clients can download the global model from the central server in REFOL, when detecting the occurrence of concept drift. 
The performance gains in REFOL demonstrate that the global model is superior to local models and the FOL paradigm is more effective than the distributed learning paradigm.
Compared with REFOL-V1, REFOL can offer performance gains of 74\% in terms of RMSE, which indicates the high efficiency of data-driven client participation mechanism.
In REFOL-V2, the central server utilizes averaging aggregation strategy to yield the updated global model without evaluating participating importance of clients.
Consequently, global model generalization in REFOL-V2 decreases and the predicting performance declines compared with REFOL, indicating the effectiveness of graph convolution-based aggregation mechanism.

\begin{figure}[!ht]
	\centering
	\includegraphics[width=0.4\textwidth]{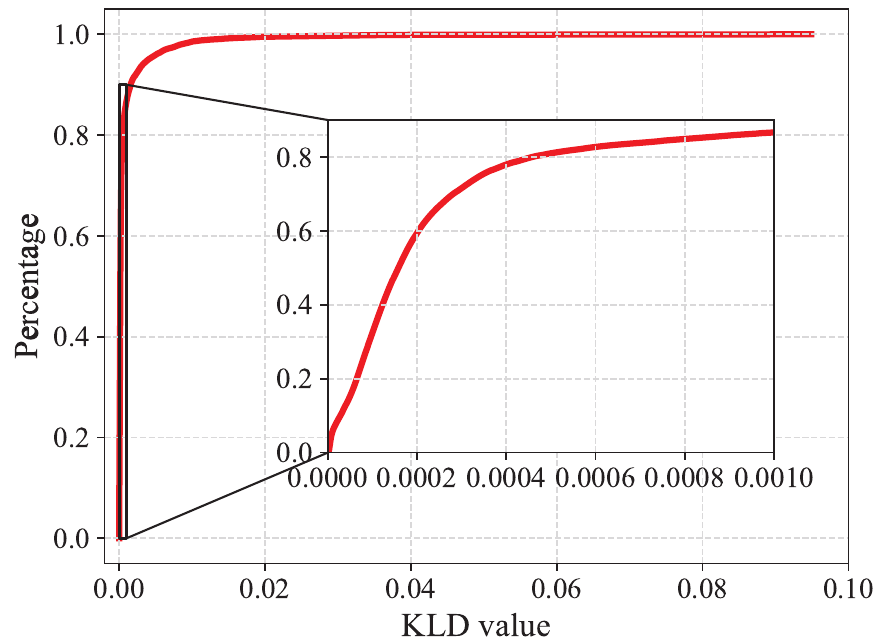}
	\caption{Distribution of KLD values on PEMS-BAY dataset.}
	\label{kl_distribution}
\end{figure}

\begin{figure*}[!ht]
	\centering
	\includegraphics[width=0.9\textwidth,height=0.5\textwidth]{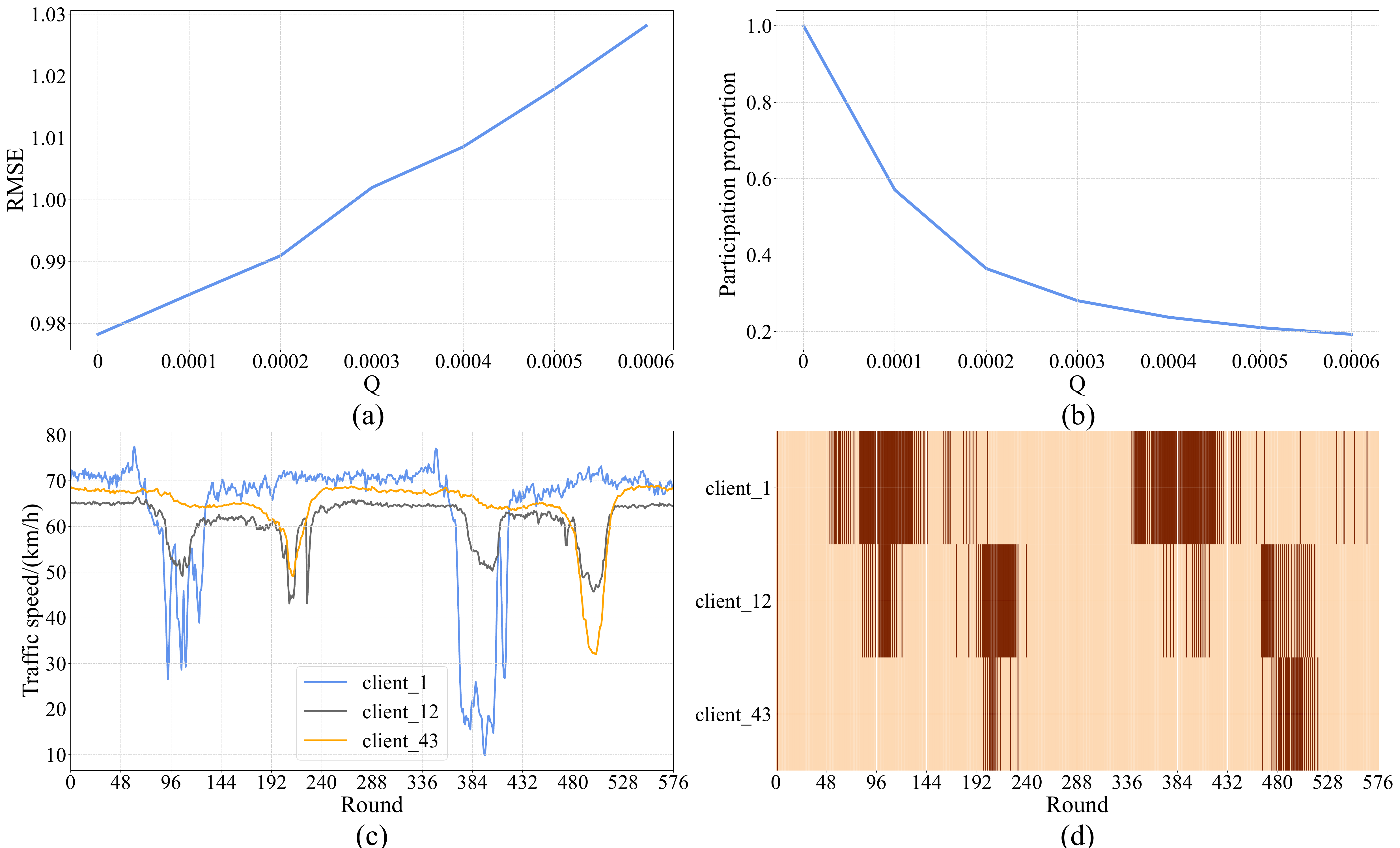}
	\caption{\textcolor{black}{(a) Prediction RMSE versus Q, (b) participation proportion versus Q, (c) traffic flows of three random clients, and (d) participation of these three clients.}}
	\label{round}
\end{figure*}

\subsection{Influence of Q on Prediction Performance}
\label{Q}
In this subsection, \textcolor{black}{we analyze how determine the value of $Q$ and explore the influence of $Q$ on prediction performance.}
We first calculate the KLD values between two neighbour traffic sequences for each client, and then calculate the cumulative distribution of these values. The results on PEMS-BAY dataset are shown in Fig. \ref{kl_distribution}. 
\textcolor{black}{It is plain that most KLD values are less than 0.001 and especially range from 0 to 0.0006. 
% With the increase of KLD values, the curve ascends slowly. That is because the drastic fluctuation in traffic flows seldom occurs. 
Then, we conduct experiments with 7 different settings of $Q$ (from 0 to 0.0006).
Fig. \ref{round} (a) and (b) show how RMSE and participation proportion change with $Q$ varying. 
We can observe that when $Q$ increases, the concept drift detection criterion is less strict and fewer client participate in federated training (participation proportion declines with $Q$ increasing), leading to poorer performance (RMSE increases with $Q$ increasing).
Specifically, the prediction performance of REFOL is consistently superior to that of the baselines, until $Q$ increases to 0.0003. When $Q=0.0003$, the participation proportion is 28\%. Therefore, the computational cost for local optimization and communication cost for exchanging model parameters can be decreased by up to 72\%.
Therefore, to strike a balance between prediction performance and resource consumption, $Q$ is finally determined as 0.0003.
}

Furthermore, we explore the efficiency of data-driven client participation mechanism. 
We randomly choose experimental results of $3$ clients with $F=1$ and $Q=0.0003$ on PEMS-BAY dataset.
Fig. \ref{round} (c) and (d) show the raw traffic flows and the participation of these three clients respectively.
In Fig. \ref{round} (d), when the client downloads the global model at a certain round, the corresponding region is marked as red.
It is intuitive that the participation of clients has much to do with the raw raw traffic flows.
Specifically, in the range of the 48-th to 144-th round, there is fluctuation in the traffic flow of $r_1$ and the KLD values are over the given KLD threshold. Therefore, $r_1$ participates and requests the global model from the central server at most rounds in this range.
However, the traffic flow of $r_{43}$ is smooth enough, and $r_{43}$ does not detect concept drift and hence reuses the local model for prediction.
Therefore, we argue that our proposed data-driven client participation mechanism is efficient in detecting concept drift, and thus enables clients to reasonably  determine whether to participate in model updates at each round.

\begin{figure}[!ht]
	\centering
	\includegraphics[width=0.48\textwidth]{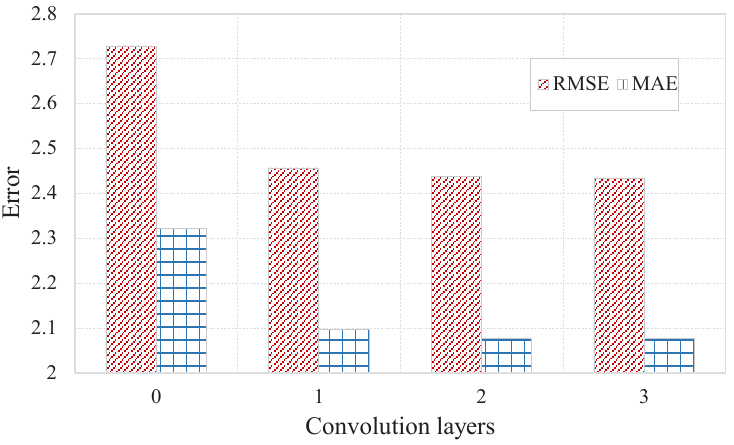}
	\caption{The number of graph convolution layers versus prediction performance on PEMS-BAY dataset.}
	\label{layer}
\end{figure}

\subsection{Effect of Graph Convolution Layers on Prediction Performance}

% We design the graph convolution-based aggregation mechanism to build the evaluation of spatial correlation in the process of aggregation, thus improving the generalization ability of the yielded global models. 
The number of graph convolution layers determines how well the central server can evaluate spatial dependence among participants and ultimately affects the prediction performance.
Hence, in this subsection, we explore the effects of graph convolution layers on prediction performance. 
The values of RMSE and MAE with different numbers of convolution layers on PEMS-BAY dataset are shown in Fig. \ref{layer}. When the number of convolution layers is $0$, the central server adopts the conventional averaging aggregation mechanism to generate the fresh global model.
It is intuitive that our proposed graph convolution-based model aggregation mechanism offers much more performance gains compared with the averaging aggregation mechanism, which demonstrates the effectiveness of our proposed aggregation mechanism.
Furthermore, with 1-layer graph convolution, the central server fails to evaluate spatial correlation comprehensively, thus yielding subpar prediction performance.
With 2-layer graph convolution, the spatial correlation among participants can be well evaluated. 
When the layer number is greater than 2, the increased layers of graph convolution will pose more computational pressure on the central server but offer few performance improvement. 
As shown in Fig. \ref{layer}, the performance difference with $2$- and $3$-layer graph convolution is negligible.
Overall, the experimental results demonstrate the superiority of our proposed graph convolution-based model aggregation mechanism. This reason is that the novel mechanism embraces the assessment of spatial correlation among participants, which can increase global model generalization and further improve prediction performance.

\section{Conclusion and Discussion}
In this paper, we investigate the \textcolor{black}{TFF} problem and propose a novel FOL method named REFOL, aiming at improving prediction performance without bringing out unnecessary computational and communication overhead.
We design a data-driven client participation mechanism to detect the occurrence of concept drift by evaluating distribution divergence of traffic data.
Accordingly, each client decides whether to participate in model updates and further performs adaptive online optimization at each round, which can not only guarantee the prediction performance but also avoid unnecessary computing and communication overhead for insignificant model optimization.
Furthermore, we build the immediate evaluation of time-varying spatial correlation in the aggregation process and propose a graph convolution-based model aggregation mechanism, which gets rid of client-side extra resource waste for evaluating spatial dependence like existing FL methods.
Finally, comprehensive experiments are conducted on PEMS-BAY and METR-LA datasets to validate the superiority of REFOL in terms of prediction improvement and resource saving.

To increase the transparency and reliability of our REFOL, we conduct further discussion as follows.

\begin{itemize}
    \item \textcolor{black}{\textbf{Specificity.} 
    REFOL is oriented for \textit{efficiently} tackling the concept drift (i.e., traffic distribution shift) for traffic forecasting tasks in FOL paradigm. Different from FedOSTC, where clients need to update local models each round, REFOL enables clients to autonomously detect concept drift and then determine whether to update local models.
    By this means, clients prefer to reuse the saved historical prediction models for prediction, which can guarantee the prediction performance and simultaneously reduce resource overhead resulting from frequent local update.
    Moreover, we incorporate the evaluation of spatial correlation into the aggregation process to decrease the communication overhead.
    Given the ubiquitous distribution drift, we analyze that REFOL can apply to other spatio-temporal forecasting tasks, such as cellular traffic forecasting, weather forecasting, and retail sales forecasting, etc. 
    In different tasks, the KLD threshold $Q$ can adjusted to satisfy specific requirements in prediction accuracy and resource consumption.
    For example, if we prefer higher prediction accuracy, $Q$ can be set to lower values, which leads to more strict concept drift criteria and clients will update local models more frequently.
    If we emphasize consumption reduction, $Q$ can be set to higher values, and therefore clients are more likely to reuse the historical model parameters, decreasing the resource consumption.}
    \item \textbf{Scalability.} The proposed REFOL can adapt to real-world traffic forecasting scenarios with hundreds or thousands of traffic nodes. 
    Given the complex network conditions in real world, the process of exchanging model parameters between nodes and the central server may be blocked. 
    Therefore, we can supplement waiting time stamps for the central server and nodes. If the time stamp at node expires, it will reuse the locally-saved model parameters. If the time stamp at the central server expires, it will aggregate the punctual model parameters. Moreover, considering that distant nodes have little or no correlation, the topological graph can be split into multiple subgraphs to mitigate the computation load of the central server. We will follow the idea to refine REFOL to accommodate to large-scale traffic nodes in the future works.
    \item \textbf{Limitations and Future Works.} We assume that all traffic nodes are furnished with similar computing and communication resources. However, traffic nodes may be equipped with imbalanced resources in real world. In such asynchronous FL paradigm, a simple strategy of adapting REFOL may be setting time stamps at the central server and traffic nodes. However, how to guarantee the prediction performance and efficiency is non-trivial. Hence, in the future works, we will dive into asynchronous FL methods for TFF.
    \end{itemize}

\section*{Acknowledgments}
% This should be a simple paragraph before the References to thank those individuals and institutions who have supported your work on this article.
This work is supported by the National Key Research and Development Program of China (2021YFB2900102) and the National Natural Science Foundation of China (No. 62072436 and No. 62202449).

% {\appendix[Proof of the Zonklar Equations]
% Use $\backslash${\tt{appendix}} if you have a single appendix:
% Do not use $\backslash${\tt{section}} anymore after $\backslash${\tt{appendix}}, only $\backslash${\tt{section*}}.
% If you have multiple appendixes use $\backslash${\tt{appendices}} then use $\backslash${\tt{section}} to start each appendix.
% You must declare a $\backslash${\tt{section}} before using any $\backslash${\tt{subsection}} or using $\backslash${\tt{label}} ($\backslash${\tt{appendices}} by itself
%  starts a section numbered zero.)}


\bibliographystyle{IEEEtran}
\bibliography{reference}

\begin{thebibliography}{10}
\providecommand{\url}[1]{#1}
\csname url@samestyle\endcsname
\providecommand{\newblock}{\relax}
\providecommand{\bibinfo}[2]{#2}
\providecommand{\BIBentrySTDinterwordspacing}{\spaceskip=0pt\relax}
\providecommand{\BIBentryALTinterwordstretchfactor}{4}
\providecommand{\BIBentryALTinterwordspacing}{\spaceskip=\fontdimen2\font plus
\BIBentryALTinterwordstretchfactor\fontdimen3\font minus
  \fontdimen4\font\relax}
\providecommand{\BIBforeignlanguage}[2]{{%
\expandafter\ifx\csname l@#1\endcsname\relax
\typeout{** WARNING: IEEEtran.bst: No hyphenation pattern has been}%
\typeout{** loaded for the language `#1'. Using the pattern for}%
\typeout{** the default language instead.}%
\else
\language=\csname l@#1\endcsname
\fi
#2}}
\providecommand{\BIBdecl}{\relax}
\BIBdecl

\bibitem{afrin2020survey}
T.~Afrin and N.~Yodo, ``A survey of road traffic congestion measures towards a
  sustainable and resilient transportation system,'' \emph{Sustainability},
  vol.~12, no.~11, p. 4660, 2020.

\bibitem{guo2019attention}
S.~Guo, Y.~Lin, N.~Feng, C.~Song, and H.~Wan, ``Attention based
  spatial-temporal graph convolutional networks for traffic flow forecasting,''
  in \emph{Proceedings of the AAAI conference on artificial intelligence},
  vol.~33, no.~01, 2019, pp. 922--929.

\bibitem{miglani2019deep}
A.~Miglani and N.~Kumar, ``Deep learning models for traffic flow prediction in
  autonomous vehicles: A review, solutions, and challenges,'' \emph{Vehicular
  Communications}, vol.~20, p. 100184, 2019.

\bibitem{8894895}
F.~Zhao, G.-Q. Zeng, and K.-D. Lu, ``Enlstm-wpeo: Short-term traffic flow
  prediction by ensemble lstm, nnct weight integration, and population extremal
  optimization,'' \emph{IEEE Transactions on Vehicular Technology}, vol.~69,
  no.~1, pp. 101--113, 2020.

\bibitem{liang2019urbanfm}
Y.~Liang, K.~Ouyang, L.~Jing, S.~Ruan, Y.~Liu, J.~Zhang, D.~S. Rosenblum, and
  Y.~Zheng, ``Urbanfm: Inferring fine-grained urban flows,'' in
  \emph{Proceedings of the 25th ACM SIGKDD international conference on
  knowledge discovery \& data mining}, 2019, pp. 3132--3142.

\bibitem{zheng2020gman}
C.~Zheng, X.~Fan, C.~Wang, and J.~Qi, ``Gman: A graph multi-attention network
  for traffic prediction,'' in \emph{Proceedings of the AAAI conference on
  artificial intelligence}, vol.~34, no.~01, 2020, pp. 1234--1241.

\bibitem{yang2019federated}
Q.~Yang, Y.~Liu, T.~Chen, and Y.~Tong, ``Federated machine learning: Concept
  and applications,'' \emph{ACM Transactions on Intelligent Systems and
  Technology (TIST)}, vol.~10, no.~2, pp. 1--19, 2019.

\bibitem{liu2020privacy}
Y.~Liu, J.~James, J.~Kang, D.~Niyato, and S.~Zhang, ``Privacy-preserving
  traffic flow prediction: A federated learning approach,'' \emph{IEEE Internet
  of Things Journal}, vol.~7, no.~8, pp. 7751--7763, 2020.

\bibitem{zhang2021fastgnn}
C.~Zhang, S.~Zhang, J.~James, and S.~Yu, ``Fastgnn: A topological information
  protected federated learning approach for traffic speed forecasting,''
  \emph{IEEE Transactions on Industrial Informatics}, vol.~17, no.~12, pp.
  8464--8474, 2021.

\bibitem{meng2021cross}
C.~Meng, S.~Rambhatla, and Y.~Liu, ``Cross-node federated graph neural network
  for spatio-temporal data modeling,'' in \emph{Proceedings of the 27th ACM
  SIGKDD Conference on Knowledge Discovery \& Data Mining}, 2021, pp.
  1202--1211.

\bibitem{lu2018learning}
J.~Lu, A.~Liu, F.~Dong, F.~Gu, J.~Gama, and G.~Zhang, ``Learning under concept
  drift: A review,'' \emph{IEEE transactions on knowledge and data
  engineering}, vol.~31, no.~12, pp. 2346--2363, 2018.

\bibitem{hoi2021online}
S.~C. Hoi, D.~Sahoo, J.~Lu, and P.~Zhao, ``Online learning: A comprehensive
  survey,'' \emph{Neurocomputing}, vol. 459, pp. 249--289, 2021.

\bibitem{10618965}
Q.~Liu, S.~Sun, M.~Liu, Y.~Wang, and B.~Gao, ``Online spatio-temporal
  correlation-based federated learning for traffic flow forecasting,''
  \emph{IEEE Transactions on Intelligent Transportation Systems}, vol.~25,
  no.~10, pp. 13\,027--13\,039, 2024.

\bibitem{wei2006time}
W.~W. Wei, ``Time series analysis,'' in \emph{The Oxford Handbook of
  Quantitative Methods in Psychology: Vol. 2}, 2006.

\bibitem{van1996combining}
M.~Van Der~Voort, M.~Dougherty, and S.~Watson, ``Combining kohonen maps with
  arima time series models to forecast traffic flow,'' \emph{Transportation
  Research Part C: Emerging Technologies}, vol.~4, no.~5, pp. 307--318, 1996.

\bibitem{lee1999application}
S.~Lee and D.~B. Fambro, ``Application of subset autoregressive integrated
  moving average model for short-term freeway traffic volume forecasting,''
  \emph{Transportation research record}, vol. 1678, no.~1, pp. 179--188, 1999.

\bibitem{yang2019traffic}
B.~Yang, S.~Sun, J.~Li, X.~Lin, and Y.~Tian, ``Traffic flow prediction using
  lstm with feature enhancement,'' \emph{Neurocomputing}, vol. 332, pp.
  320--327, 2019.

\bibitem{dai2019short}
G.~Dai, C.~Ma, and X.~Xu, ``Short-term traffic flow prediction method for urban
  road sections based on space--time analysis and gru,'' \emph{IEEE Access},
  vol.~7, pp. 143\,025--143\,035, 2019.

\bibitem{sun2020ssgru}
P.~Sun, A.~Boukerche, and Y.~Tao, ``Ssgru: A novel hybrid stacked gru-based
  traffic volume prediction approach in a road network,'' \emph{Computer
  Communications}, vol. 160, pp. 502--511, 2020.

\bibitem{hsueh2021short}
Y.-L. Hsueh and Y.-R. Yang, ``A short-term traffic speed prediction model based
  on lstm networks,'' \emph{International journal of intelligent transportation
  systems research}, vol.~19, no.~3, pp. 510--524, 2021.

\bibitem{li2018diffusion}
\BIBentryALTinterwordspacing
Y.~Li, R.~Yu, C.~Shahabi, and Y.~Liu, ``Diffusion convolutional recurrent
  neural network: Data-driven traffic forecasting,'' in \emph{International
  Conference on Learning Representations}, 2018. [Online]. Available:
  \url{https://openreview.net/forum?id=SJiHXGWAZ}
\BIBentrySTDinterwordspacing

\bibitem{zhang2024personalized}
Y.~Zhang, H.~Lu, N.~Liu, Y.~Xu, Q.~Li, and L.~Cui, ``Personalized federated
  learning for cross-city traffic prediction,'' in \emph{33rd International
  Joint Conference on Artificial Intelligence, IJCAI 2024}.\hskip 1em plus
  0.5em minus 0.4em\relax International Joint Conferences on Artificial
  Intelligence, 2024, pp. 5526--5534.

\bibitem{xia2022short}
M.~Xia, D.~Jin, and J.~Chen, ``Short-term traffic flow prediction based on
  graph convolutional networks and federated learning,'' \emph{IEEE
  Transactions on Intelligent Transportation Systems}, 2022.

\bibitem{liu2023multilevel}
L.~Liu, Y.~Tian, C.~Chakraborty, J.~Feng, Q.~Pei, L.~Zhen, and K.~Yu,
  ``Multilevel federated learning-based intelligent traffic flow forecasting
  for transportation network management,'' \emph{IEEE Transactions on Network
  and Service Management}, vol.~20, no.~2, pp. 1446--1458, 2023.

\bibitem{lou2022stfl}
G.~Lou, Y.~Liu, T.~Zhang, and X.~Zheng, ``Stfl: A spatial-temporal federated
  learning framework for graph neural networks,'' in \emph{AAAI Conference on
  Artificial Intelligence Workshop on Deep Learning on Graphs: Methods and
  Applications}, 2022.

\bibitem{yuan2022fedstn}
X.~Yuan, J.~Chen, J.~Yang, N.~Zhang, T.~Yang, T.~Han, and A.~Taherkordi,
  ``Fedstn: Graph representation driven federated learning for edge computing
  enabled urban traffic flow prediction,'' \emph{IEEE Transactions on
  Intelligent Transportation Systems}, 2022.

\bibitem{yang2024fedgtp}
L.~Yang, W.~Chen, X.~He, S.~Wei, Y.~Xu, Z.~Zhou, and Y.~Tong, ``Fedgtp:
  Exploiting inter-client spatial dependency in federated graph-based traffic
  prediction,'' in \emph{Proceedings of the 30th ACM SIGKDD Conference on
  Knowledge Discovery and Data Mining}, 2024, pp. 6105--6116.

\bibitem{lu2016concept}
N.~Lu, J.~Lu, G.~Zhang, and R.~L. De~Mantaras, ``A concept drift-tolerant
  case-base editing technique,'' \emph{Artificial Intelligence}, vol. 230, pp.
  108--133, 2016.

\bibitem{kifer2004detecting}
D.~Kifer, S.~Ben-David, and J.~Gehrke, ``Detecting change in data streams,'' in
  \emph{VLDB}, vol.~4.\hskip 1em plus 0.5em minus 0.4em\relax Toronto, Canada,
  2004, pp. 180--191.

\bibitem{gu2016concept}
F.~Gu, G.~Zhang, J.~Lu, and C.-T. Lin, ``Concept drift detection based on equal
  density estimation,'' in \emph{2016 International Joint Conference on Neural
  Networks (IJCNN)}.\hskip 1em plus 0.5em minus 0.4em\relax IEEE, 2016, pp.
  24--30.

\bibitem{bu2017incremental}
L.~Bu, D.~Zhao, and C.~Alippi, ``An incremental change detection test based on
  density difference estimation,'' \emph{IEEE Transactions on Systems, Man, and
  Cybernetics: Systems}, vol.~47, no.~10, pp. 2714--2726, 2017.

\bibitem{gama2004learning}
J.~Gama, P.~Medas, G.~Castillo, and P.~Rodrigues, ``Learning with drift
  detection,'' in \emph{Advances in Artificial Intelligence--SBIA 2004: 17th
  Brazilian Symposium on Artificial Intelligence, Sao Luis, Maranhao, Brazil,
  September 29-Ocotber 1, 2004. Proceedings 17}.\hskip 1em plus 0.5em minus
  0.4em\relax Springer, 2004, pp. 286--295.

\bibitem{baena2006early}
M.~Baena-Garc{\i}a, J.~del Campo-{\'A}vila, R.~Fidalgo, A.~Bifet, R.~Gavalda,
  and R.~Morales-Bueno, ``Early drift detection method,'' in \emph{Fourth
  international workshop on knowledge discovery from data streams},
  vol.~6.\hskip 1em plus 0.5em minus 0.4em\relax Citeseer, 2006, pp. 77--86.

\bibitem{nishida2007detecting}
K.~Nishida and K.~Yamauchi, ``Detecting concept drift using statistical
  testing,'' in \emph{Discovery science}, vol. 4755.\hskip 1em plus 0.5em minus
  0.4em\relax Springer, 2007, pp. 264--269.

\bibitem{mcmahan2017communication}
B.~McMahan, E.~Moore, D.~Ramage, S.~Hampson, and B.~A. y~Arcas,
  ``Communication-efficient learning of deep networks from decentralized
  data,'' in \emph{Artificial intelligence and statistics}.\hskip 1em plus
  0.5em minus 0.4em\relax PMLR, 2017, pp. 1273--1282.

\bibitem{joyce2011kullback}
J.~M. Joyce, ``Kullback-leibler divergence,'' in \emph{International
  encyclopedia of statistical science}.\hskip 1em plus 0.5em minus 0.4em\relax
  Springer, 2011, pp. 720--722.

\bibitem{cho-etal-2014-learning}
K.~Cho, B.~van Merri{\"e}nboer, C.~Gulcehre, D.~Bahdanau, F.~Bougares,
  H.~Schwenk, and Y.~Bengio, ``Learning phrase representations using {RNN}
  encoder{--}decoder for statistical machine translation,'' in
  \emph{Proceedings of the 2014 Conference on Empirical Methods in Natural
  Language Processing ({EMNLP})}, Doha, Qatar, Oct. 2014.

\bibitem{kipfsemi}
T.~N. Kipf and M.~Welling, ``Semi-supervised classification with graph
  convolutional networks,'' in \emph{International Conference on Learning
  Representations, 2017}, 2017.

\bibitem{feng2006svm}
H.~Feng, Y.~Shu, S.~Wang, and M.~Ma, ``Svm-based models for predicting wlan
  traffic,'' in \emph{2006 IEEE international conference on communications},
  vol.~2.\hskip 1em plus 0.5em minus 0.4em\relax IEEE, 2006, pp. 597--602.

\bibitem{yu2018spatio}
B.~Yu, H.~Yin, and Z.~Zhu, ``Spatio-temporal graph convolutional networks: a
  deep learning framework for traffic forecasting,'' in \emph{27th
  International Joint Conference on Artificial Intelligence, IJCAI 2018}, 2018,
  pp. 3634--3640.

\bibitem{jiang2023spatio}
R.~Jiang, Z.~Wang, J.~Yong, P.~Jeph, Q.~Chen, Y.~Kobayashi, X.~Song,
  S.~Fukushima, and T.~Suzumura, ``Spatio-temporal meta-graph learning for
  traffic forecasting,'' in \emph{Proceedings of the AAAI conference on
  artificial intelligence}, vol.~37, no.~7, 2023, pp. 8078--8086.

\bibitem{gneiting2007strictly}
T.~Gneiting and A.~E. Raftery, ``Strictly proper scoring rules, prediction, and
  estimation,'' \emph{Journal of the American statistical Association}, vol.
  102, no. 477, pp. 359--378, 2007.

\bibitem{SPILIOTIS2020550}
\BIBentryALTinterwordspacing
E.~Spiliotis, V.~Assimakopoulos, and S.~Makridakis, ``Generalizing the theta
  method for automatic forecasting,'' \emph{European Journal of Operational
  Research}, vol. 284, no.~2, pp. 550--558, 2020. [Online]. Available:
  \url{https://www.sciencedirect.com/science/article/pii/S0377221720300242}
\BIBentrySTDinterwordspacing

\bibitem{molchanov2019pruning}
P.~Molchanov, S.~Tyree, T.~Karras, T.~Aila, and J.~Kautz, ``Pruning
  convolutional neural networks for resource efficient inference,'' in
  \emph{5th International Conference on Learning Representations, ICLR
  2017-Conference Track Proceedings}, 2019.

\bibitem{epoch2021backwardforwardFLOPratio}
\BIBentryALTinterwordspacing
M.~Hobbhahn and J.~Sevilla, ``What’s the backward-forward flop ratio for
  neural networks?'' 2021, accessed: 2023-4-18. [Online]. Available:
  \url{https://epochai.org/blog/backward-forward-FLOP-ratio}
\BIBentrySTDinterwordspacing

\end{thebibliography}

\newpage

\begin{IEEEbiography}[{\includegraphics[width=1in,height=1.25in,clip,keepaspectratio]{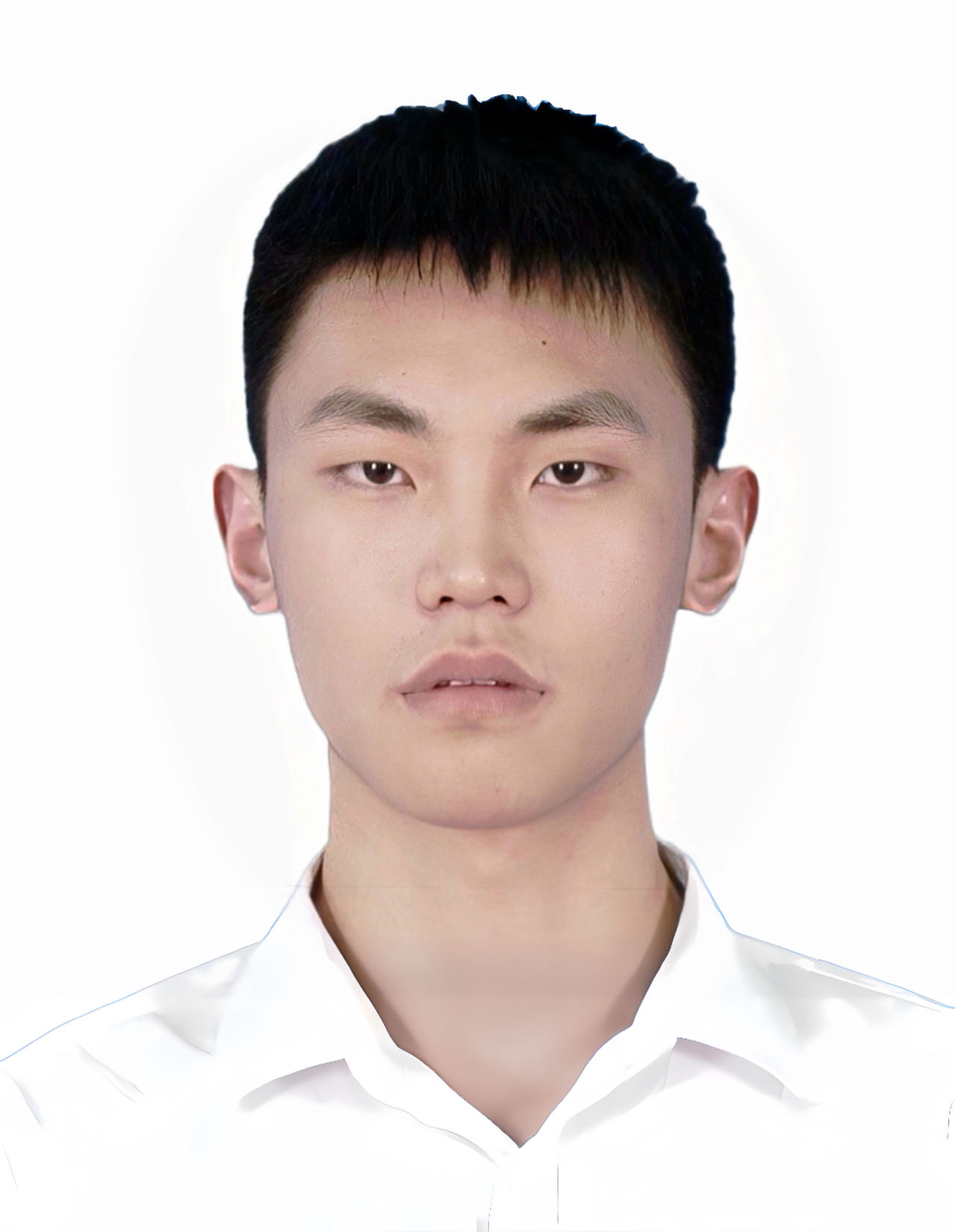}}]{Qingxiang Liu}
	is currently pursuing the Ph.D. degree with the Institute of Computing Technology, Chinese Academy of Sciences, Beijing, China. 
	He has authored or coauthored several papers at IEEE Transactions on Intelligent Transportation Systems, Future Generation Computer Systems and so on. His current research interests include federated learning , time series forecasting and spatio-temporal mining. 
\end{IEEEbiography}

\vspace{-1cm}
\begin{IEEEbiography}[{\includegraphics[width=1in,height=1.25in,clip,keepaspectratio]{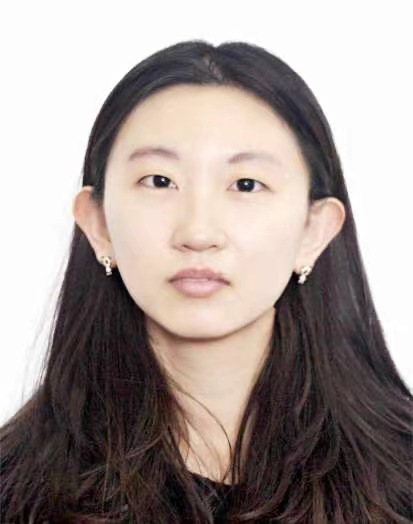}}]{Sheng Sun}
 received her B.S. and Ph.D degrees in computer science from Beihang University, China, and the University of Chinese Academy of Sciences, China, respectively. She is currently an assistant professor at the Institute of Computing Technology, Chinese Academy of Sciences, Beijing, China. Her current research interests include federated learning, mobile computing and edge intelligence. 
\end{IEEEbiography}

\vspace{-1cm}

\begin{IEEEbiography}[{\includegraphics[width=1in,height=1.25in,clip,keepaspectratio]{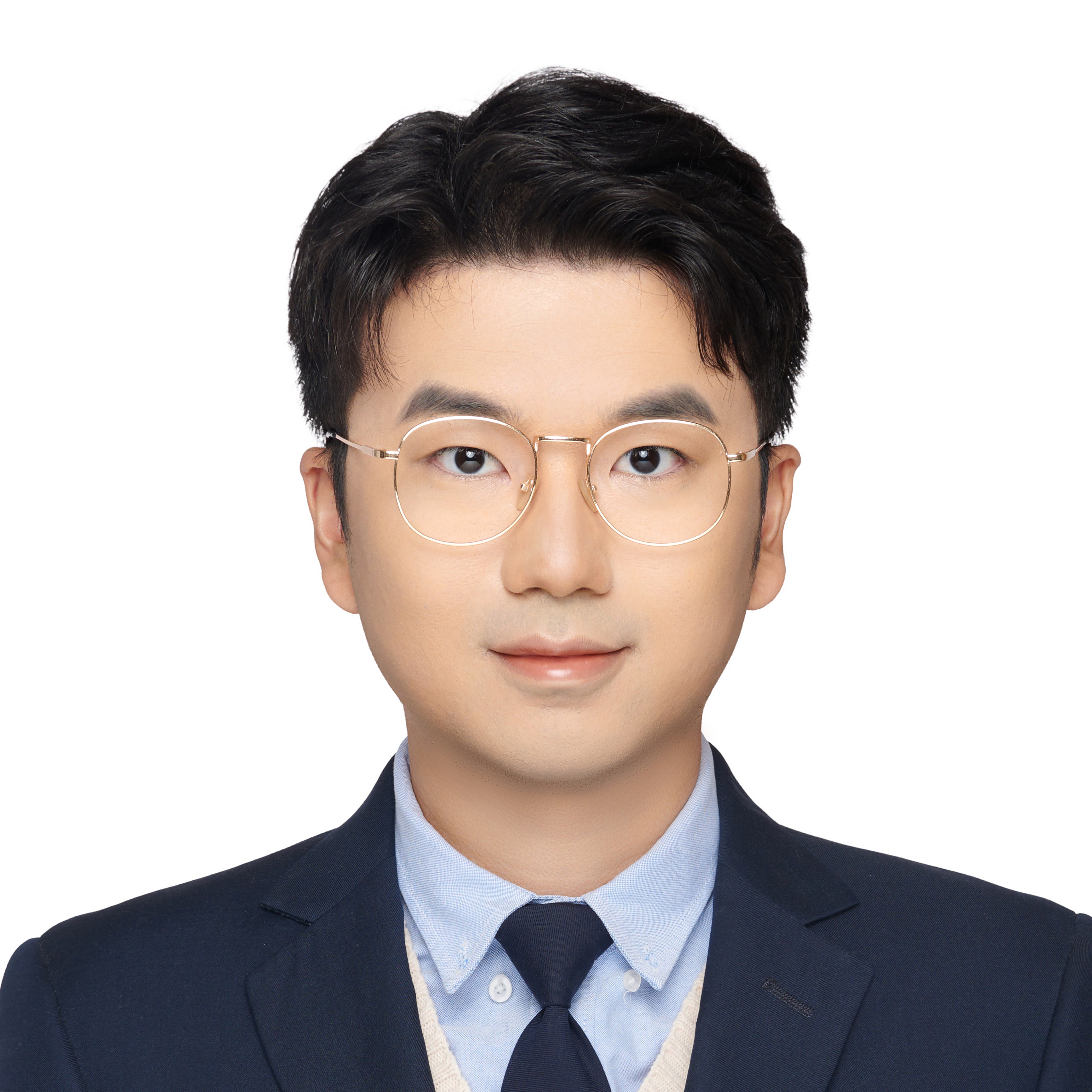}}]{Yuxuan Liang}
is currently a tenure-track Assistant Professor at Hong Kong University of Science and Technology (Guangzhou), working on the research, development, and innovation of spatio-temporal data mining and AI, with a broad range of applications in urban computing. Prior to that, he obtained his PhD degree at the School of Computing, National University of Singapore. He has published over 70 papers in refereed journals (e.g., TPAMI, AI, TKDE) and conferences (e.g., KDD, NeurIPS, ICML, ICLR, WWW). These publications have attracted 3,700 citations with h-index of 30. He has served as a program committee member for prestigious conferences, such as KDD, ICML, ICLR, NeurIPS, WWW, CVPR, and ICCV. He has served as organizer or co-chair of Workshop on Urban Computing (Urbcomp@KDD-23) and Workshop on AI for Time Series (AI4TS@IJCAI-24). He has received The 23rd China Patent Excellence Award and SDSC Dissertation Research Fellowship 2020.
\end{IEEEbiography}

\vspace{-1cm}

\begin{IEEEbiography}[{\includegraphics[width=1in,height=1.25in,clip,keepaspectratio]{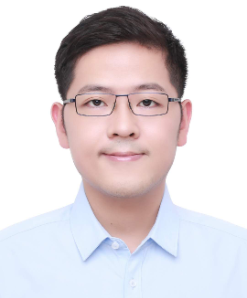}}]{Xiaolong Xu}
(Senior Member, IEEE) received the Ph.D. degree in computer science and technology from Nanjing University, China, in 2016. He is currently a Full Professor with the School of Software, Nanjing University of Information Science and Technology. He received the Best Paper Award from the IEEE CBD 2016, IEEE CyberTech2021, CENET2021, IEEE iThings2022 and EAI CloudComp2022, and the Outstanding Paper Award from IEEE SmartCity2021. He also received the Annual Best Paper Award from Elsevier JNCA. He has been selected as the Highly Cited Researcher of Clarivate 2021 and 2022. His research interests include edge intelligence and service computing.
\end{IEEEbiography}

\vspace{-1cm}

\begin{IEEEbiography}[{\includegraphics[width=1in,height=1.25in,clip,keepaspectratio]{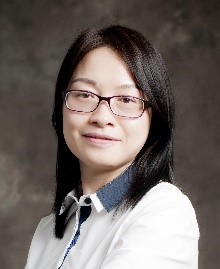}}]{Min Liu}
(Senior Member, IEEE) received her B.S. and M.S. degrees in computer science from Xi’an Jiaotong University, China, in 1999 and 2002, respectively. She got her Ph.D in computer science from the Graduate University of the Chinese Academy of Sciences in 2008. She is currently a professor at the Networking Technology Research Centre, Institute of Computing Technology, Chinese Academy of Sciences. Her current research interests include mobile computing and edge intelligence.
\end{IEEEbiography}

\vspace{-1cm}

\begin{IEEEbiography}[{\includegraphics[width=1in,height=1.25in,clip,keepaspectratio]{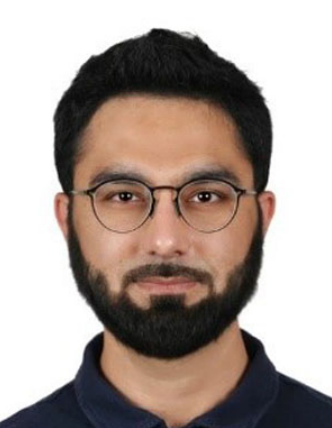}}]{Muhammad Bilal}
(Senior Member, IEEE)  received the Ph.D. degree in information and communication network engineering from the School of Electronics and Telecommunications Research Institute (ETRI), Korea University of Science and Technology, Daejeon, South Korea, in 2017. From 2018 to 2023, he was an Assistant Professor with the Division of Computer and Electronic Systems Engineering, Hankuk University of Foreign Studies, Yongin, South Korea. In 2023, he joined Lancaster University, Lancaster LA1 4YW, United Kingdom, where he is currently working as a Senior Lecturer with School of Computing and Communications. His research interests include design and analysis of network protocols, network architecture, network security, the IoT, named data networking, blockchain, cryptology, and future Internet.
\end{IEEEbiography}

\vspace{-1cm}

\begin{IEEEbiography}[{\includegraphics[width=1in,height=1.25in,clip,keepaspectratio]{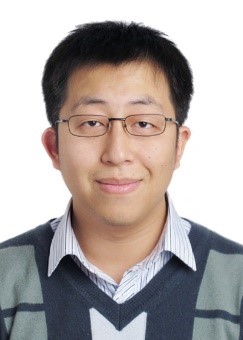}}]{Yuwei Wang}
  received his Ph.D. degree in computer science from the University of Chinese Academy of Sciences, Beijing, China. He is currently a Senior Engineer (equivalent to Associate Professor) at the Institute of Computing Technology, Chinese Academy of Sciences, Beijing, China. His current research interests include federated learning, mobile edge computing, and next-generation network architecture.
\end{IEEEbiography}
\vspace{-1cm}

\begin{IEEEbiography}[{\includegraphics[width=1in,height=1.25in,clip,keepaspectratio]{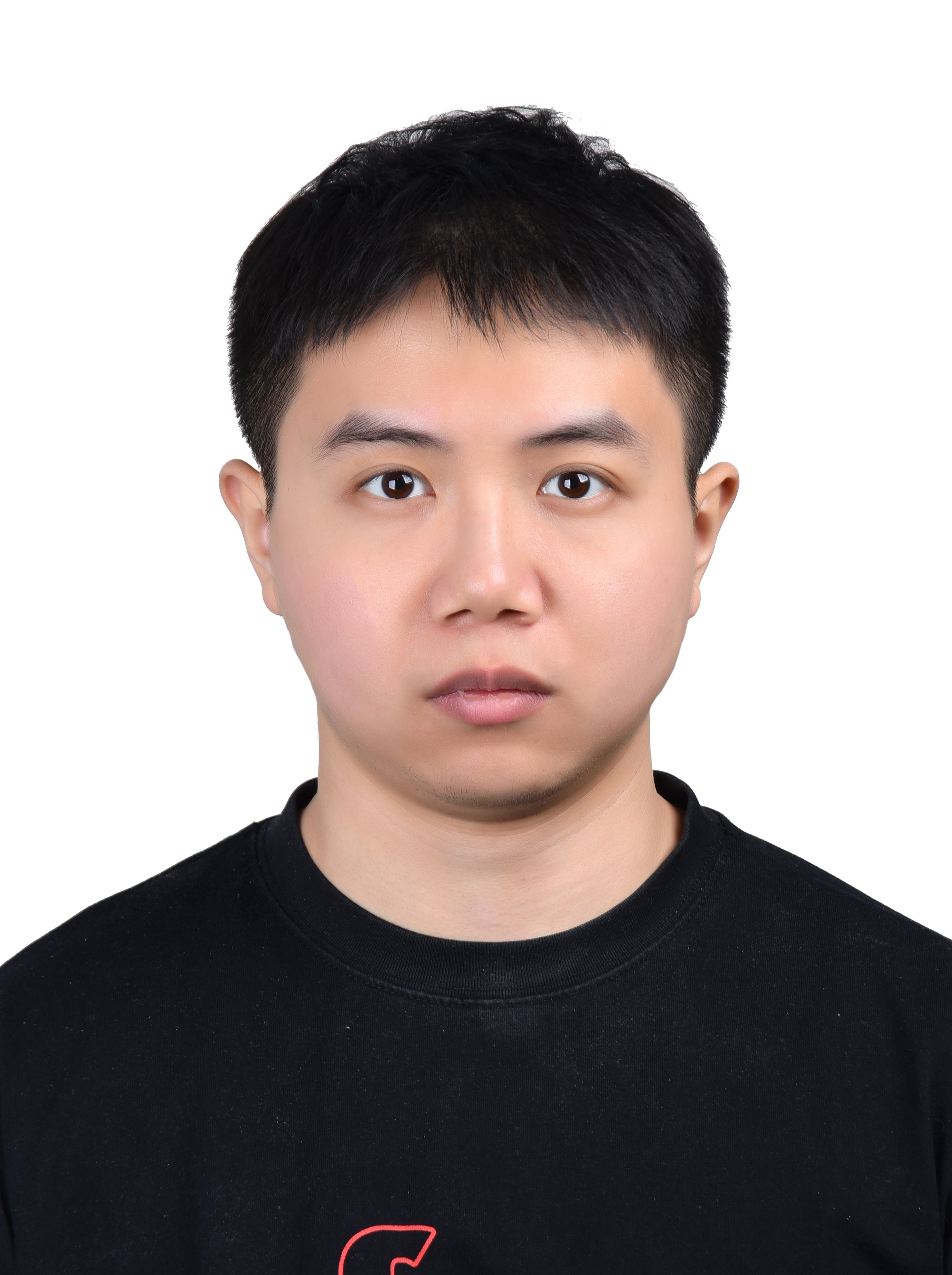}}]{Xujing Li}
 is a Ph.D. candidate with Institute of Computing Technology, Chinese Academy of Sciences, Beijing, China. His primary research focuses on Efficient Federated Learning and Distributed Learning Optimization. He received his bachelor’s degree at Central South University, Changsha, China in 2019.
\end{IEEEbiography}

\vspace{-1cm}

\begin{IEEEbiography}[{\includegraphics[width=1in,height=1.25in,clip,keepaspectratio]{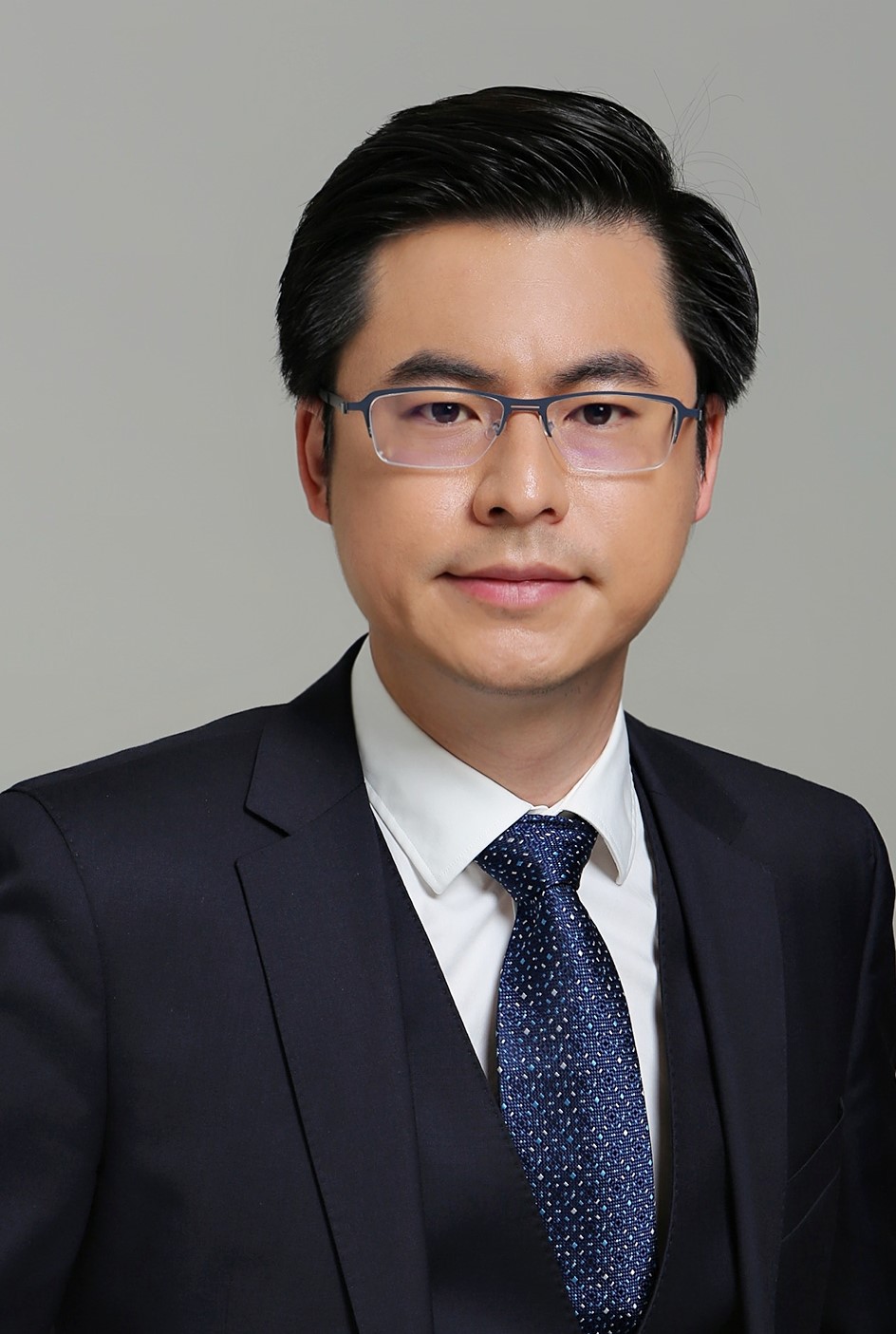}}]{Yu Zheng} 
(Fellow, IEEE) is the Vice President of JD.COM and head JD Intelligent Cities Research. Before Joining JD.COM, he was a senior research manager at Microsoft Research. He currently serves as the Editor-in-Chief of ACM Transactions on Intelligent Systems and Technology and has served as the program co-chair of ICDE 2014 (Industrial Track), CIKM 2017 (Industrial Track) and IJCAI 2019 (industrial track). He is also a keynote speaker of AAAI 2019, KDD 2019 Plenary Keynote Panel and IJCAI 2019 Industrial Days. His monograph, entitled Urban Computing, has been used as the first text book in this field. In 2013, he was named one of the Top Innovators under 35 by MIT Technology Review (TR35) and featured by Time Magazine for his research on urban computing. In 2016, Zheng was named an ACM Distinguished Scientist and elevated to an IEEE Fellow in 2020 for his contributions to spatio-temporal data mining and urban computing.
\end{IEEEbiography}

\vfill

\end{document}